\pdfoutput=1

\PassOptionsToPackage{prologue,dvipsnames}{xcolor}
\documentclass[11pt]{article}

\usepackage[preprint]{acl}

\usepackage{times}
\usepackage{latexsym}

\usepackage[T1]{fontenc}

\usepackage[utf8]{inputenc}

\usepackage{microtype}

\usepackage{inconsolata}

\usepackage{multirow}
\usepackage{booktabs}
\usepackage{tcolorbox}
\usepackage{subcaption}
\usepackage{colortbl}
\definecolor{gray}{gray}{0.85}
\tcbuselibrary{most}
\usepackage{xspace}
\usepackage{balance}
\usepackage{bm}
\usepackage{amsmath}
\usepackage{amssymb}
\usepackage{mathtools}
\usepackage{amsthm}
\usepackage{algorithm}
\usepackage{algorithmic}

\definecolor{aliceblue}{rgb}{0.94, 0.97, 1.0}

\title{Advancing General Multimodal Capability of Vision-language Models with Pyramid-descent Visual Position Encoding}

\author{Zhanpeng Chen$^{1,2}$, Mingxiao Li$^2$, Ziyang Chen$^2$, Nan Du$^2$, Xiaolong Li$^2$, Yuexian Zou$^{1,}$\thanks{Corresponding author.} \\
 $^1$Guangdong Provincial Key Laboratory of Ultra High Definition Immersive Media Technology, \\
 Shenzhen Graduate School, Peking University \\
 $^2$Tencent Hunyuan \\
  troychen927@stu.pku.edu.cn, zouyx@pku.edu.cn}

\begin{document}
\maketitle
\begin{abstract}
Vision-language Models (VLMs) have shown remarkable capabilities in advancing general artificial intelligence, yet the irrational encoding of visual positions persists in inhibiting the models' comprehensive perception performance across different levels of granularity. In this work, we propose Pyramid-descent Visual Position Encoding (PyPE), a novel approach designed to enhance the perception of visual tokens within VLMs. By assigning visual position indexes from the periphery to the center and expanding the central receptive field incrementally, PyPE addresses the limitations of traditional raster-scan methods and mitigates the long-term decay effects induced by Rotary Position Embedding (RoPE). Our method reduces the relative distance between interrelated visual elements and instruction tokens, promoting a more rational allocation of attention weights and allowing for a multi-granularity perception of visual elements and countering the over-reliance on anchor tokens. Extensive experimental evaluations demonstrate that PyPE consistently improves the general capabilities of VLMs across various sizes. Code is available at 
\url{https://github.com/SakuraTroyChen/PyPE}.
\end{abstract}

\begin{figure*}[htbp]
  \centering 
    \includegraphics[width=\linewidth]{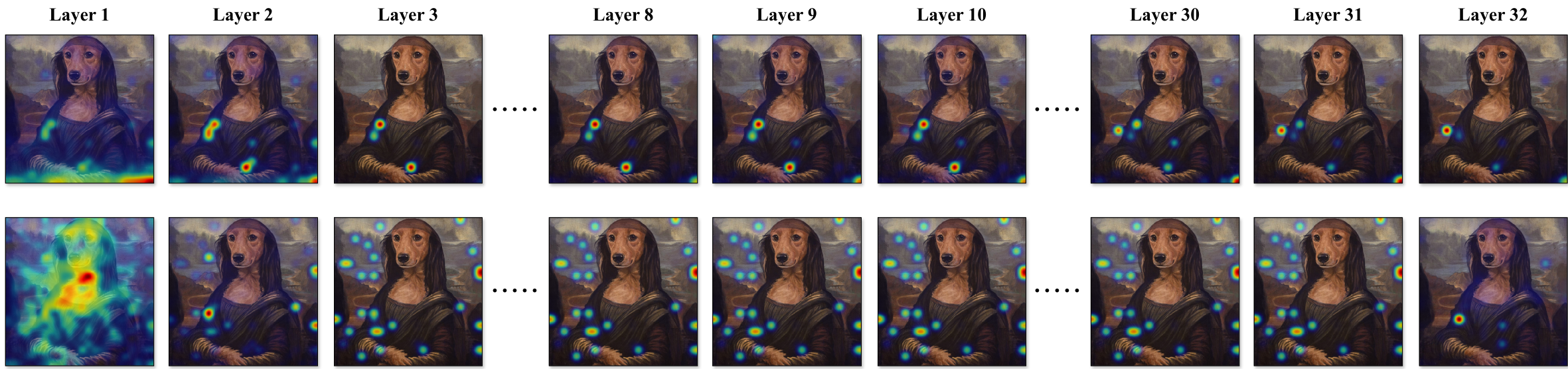}
\caption{Layer-wise attention visualization of visual-to-instruction information flow. Displayed from top to bottom are the attention heatmaps from LLaVA-1.5-7B trained with raster-scan and concentric PE, respectively. The example is derived from LLaVA-Bench~\cite{liu2024visual} and the query is \textit{"Describe this photo in detail"}.}
\label{fig:motivation} 
\end{figure*}


\section{Introduction}
Large Language Models (LLMs)~\cite{touvron2023llama,brown2020language} demonstrate significant universal capabilities that contribute to the pursuit of general artificial intelligence. However, language constitutes only one aspect of communication. Visual information plays a crucial role in augmenting and enhancing our understanding of the world. Consequently, there is a growing interest in the development of Vision-language Models (VLMs)~\cite{chen2024internvl,peng2023kosmos,wang2024qwen2,Qwen-VL} that can process and integrate visual modality. To effectively leverage the powerful contextual understanding capabilities of LLMs, VLMs project visual information to the same dimensionality as textual embeddings through specific projection layers~\cite{chen2023llava,li2023blip,zhou2024tinyllava}, which are then directly embedded into the text sequence to form the input for the foundation LLMs, enabling cross-modal alignment and instruction-following learning using next-token prediction. 

Despite their commendable progress, the typical processing of visual information does not align with the distribution patterns of visual elements. Since visual information is composed of fixed-sized patches obtained through raster scanning, patches located closer to the bottom right corner of the image are positioned nearer to the instruction tokens within the sequence. Due to the long-term decay from Rotary Position Embedding (RoPE)~\cite{su2024roformer}, visual tokens closer to the instruction tokens will be more likely to receive higher attention weights, and vice versa. This is counterintuitive, as the importance of visual information is not defined by the order of raster-scanning. \citet{xing2024mitigating} observe a similar phenomenon by visualizing the attention information flow from instruction tokens to visual tokens in the first layer of the decoder. Consequently, they propose Concentric Causal Attention (CCA), which starts assigning the position indexes of images from the peripheral and ends in the center, to alleviate the long-term decay in RoPE and improve causal attention following 2D spatial locality of images. Although CCA is both intuitive and effective, its applicability is constrained by the assumption that all significant elements related to the instructions are situated at the center of the image. This assumption inherently results in a loss of detail, limiting its effectiveness in capturing comprehensive information. 

To further investigate the impact of raster-scan and concentric PE on the fine-grained modeling of visual information, we extend the visualization to all layers of the decoder. As illustrated in Figure~\ref{fig:motivation}, CCA demonstrates exceptional performance in the first layer, alleviating the long-term decay caused by RoPE in the raster-scan approach, thereby directing the model's attention to more significant areas. However, in the subsequent layers, both methods largely maintain the same attention patterns as observed in their respective third layers, with changes only occurring in the final layer. A similar phenomenon, namely "aggregation pattern", is observed in OPERA~\cite{huang2024opera}, where both LLMs and VLMs tend to generate new tokens by concentrating on a limited number of summary tokens (also referred to as anchor tokens~\cite{wang2023label}) rather than considering all preceding tokens. This tendency towards partial overtrust leads to the neglect of fine-grained image tokens, resulting in the generation that may be hallucinatory and do not accurately reflect the image content. Moreover, it has been demonstrated in OPERA that more hallucinations are generated when more summary tokens appear in the context.

To this end, we present \textbf{Pyramid-descent Visual Position Encoding (PyPE)}, a novel position assignment approach for visual tokens, to alleviate the long-term decay induced by RoPE, avoid the "aggregation pattern" in the LLM, and ensure a comprehensive understanding of visual contents. PyPE reorganizes the flattened visual tokens into the 2D shape and assigns visual position indexes from the periphery to the center. This reduces the relative distance between interrelated visual elements, as well as the distance between significant visual elements and instruction tokens, thereby ensuring a more rational allocation of attention weights. Furthermore, to mitigate the impact of anchor tokens on the model's fine-grained perception of visual elements, we draw inspiration from Pyramid Vision Transformer (PVT)~\cite{wang2021pyramid}: consistently combining global and local receptive fields. PyPE gradually expands the central receptive field, \textit{i.e.}, the central region of the position index matrix, at predetermined intervals of layers. Specifically, we expand the central region of the position index matrix by a circle every certain number of layers. Such expansion weakens the anchor tokens and enhances the model's ability to perceive visual elements at varying levels of granularity (more cases can be found in Section~\ref{sec:case}).

With extensive experiments on visual question answering and general multimodal benchmarks, PyPE consistently improves general perception capabilities across VLMs of different sizes. In a nutshell, the main contributions of this work are as follows: 
(I) We make an in-depth analysis of how position encoding affects visual perception in VLMs. (II) Our proposed PyPE effectively mitigates long-term decay and the "aggregation pattern", which helps better perceive visual elements at different granularities. (III) Extensive evaluations demonstrate the superior performance of PyPE, a simple yet effective method that applies to any VLMs.

\begin{figure*}[htbp]
  \centering 
\subfloat[Raster-scan PE.]{
    \includegraphics[width=.3\linewidth]{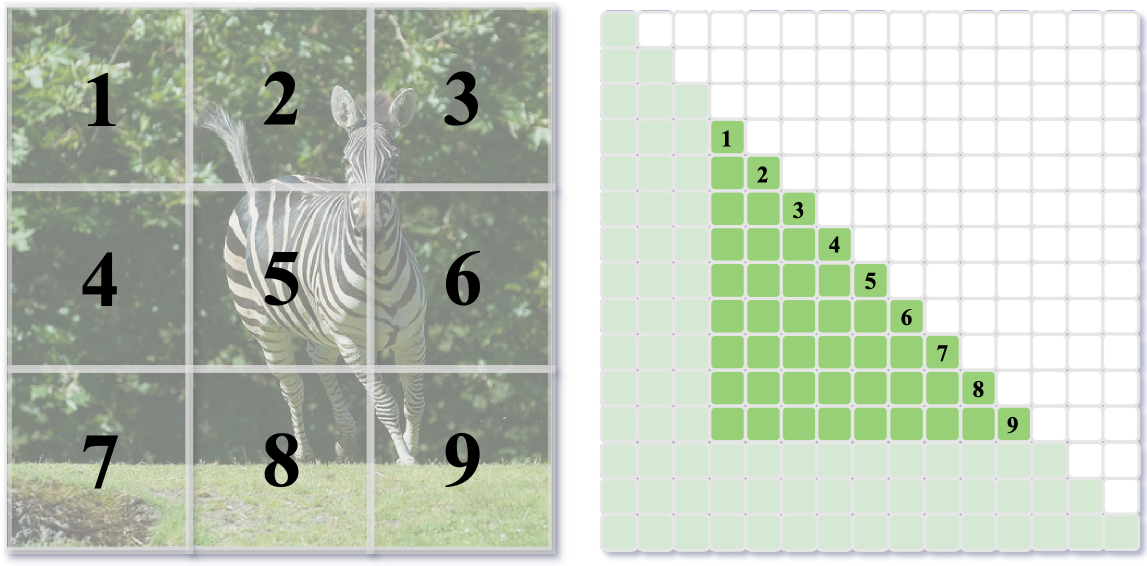}
  }
  \hfill
\subfloat[Concentric PE.]{
    \includegraphics[width=.3\linewidth]{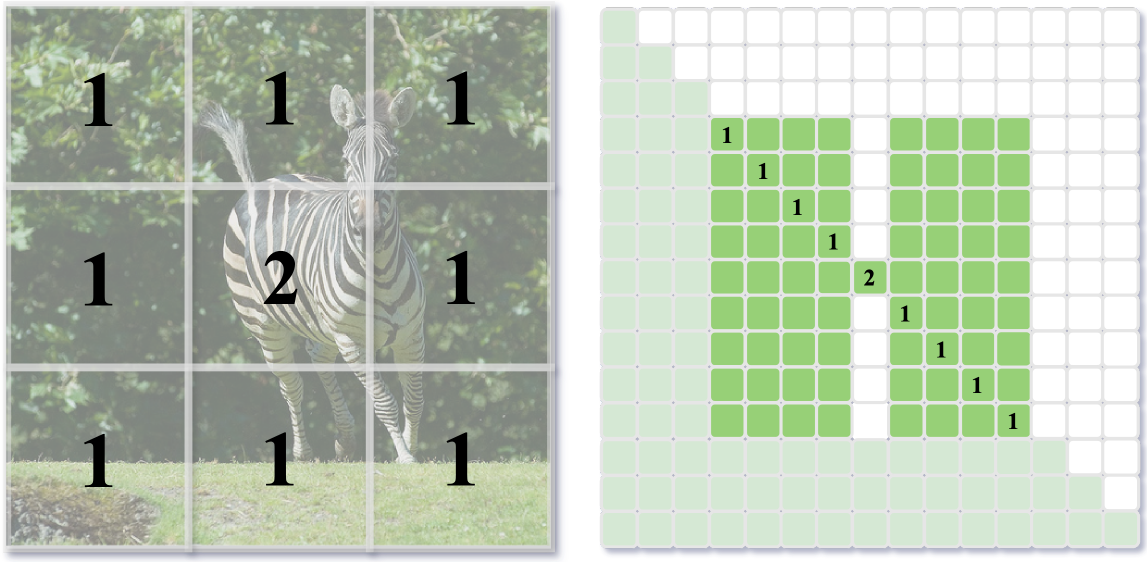}
}
  \hfill
\subfloat[All-One PE.]{
    \includegraphics[width=.3\linewidth]{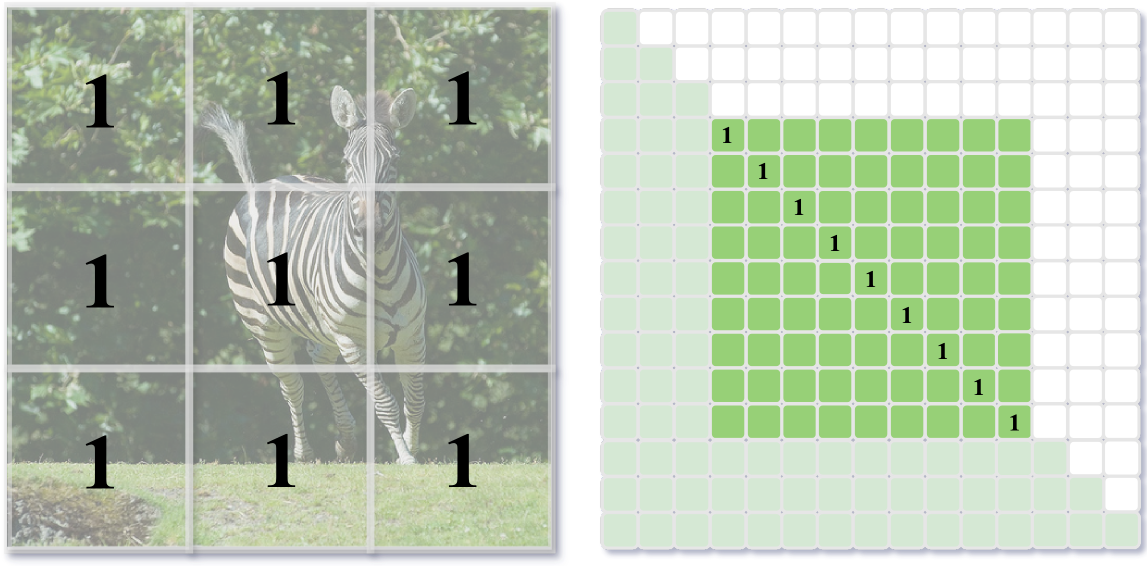}
}
\caption{An overview of patch indexes and corresponding causal mask from raster-scan, concentric, and All-One position encoding on an example from COCO~\cite{lin2014microsoft}.} 
\label{fig:PE_egs} 
\end{figure*}

\section{Related Work}
\subsection{Vision-language Model}
Recent advancements in VLMs have demonstrated impressive performance in processing multi-format information~\cite{huang2023language,achiam2023gpt}. VLMs are typically built upon existing LLMs and incorporate visual information as input tokens by utilizing an additional vision encoder (e.g., CLIP) and a bridging connector (e.g., MLP). For instance, LLaVA~\cite{liu2024improved} employs an MLP to project visual tokens and aligns the feature dimensions with word embeddings, while BLIP-2~\cite{li2023blip} utilizes a set of learnable query tokens to extract information in a query-based manner. Building upon these foundational works, MM1~\cite{mckinzie2025mm1} has further investigated the significance of the number of visual tokens and image resolution, identifying them as the most critical factors, while finding that the type of connector has minimal impact. By effectively connecting visual and textual modalities, VLMs significantly enhance human-AI interaction and exhibit remarkable capabilities in understanding and generating multimodal content~\cite{chen2024mllm,peng2023kosmos,chen2023llava,wang2024qwen2,hu2024minicpm,xie24c_interspeech}.

\subsection{Position Encoding for Transformers}
Since transformer-based models contain no recurrence~\cite{hochreiter1997long} and convolution~\cite{islam2020much} structure, additional information about the relative or absolute position of the tokens in the input sequence is required. Therefore, the community has witnessed the development of various position encoding methods, e.g. sinusoidal~\cite{vaswani2017attention}, learnable~\cite{dosovitskiy2020image}, relative~\cite{he2020deberta,shaw2018self}, and conditional~\cite{chu2021conditional} position encoding. Among these studies, RoPE~\cite{su2024roformer} is introduced to encode absolute and relative positional information, showing superiority in LLMs~\cite{touvron2023llama,achiam2023gpt}. The success of LLMs has led to the continued adoption of the effective RoPE scheme in VLMs for the unified encoding of positional information across sequences that incorporate multimodal features. However, it is important to note that visual information does not conform to the same sampling paradigm as language. The raster scanning is insufficient for modeling the spatial correlations among different patches. Consequently, numerous recent studies~\cite{chu2024visionllama,xing2024mitigating,lu2024fit} have sought to explore improved solutions that extend RoPE to visual tasks. In this paper, we investigate a novel multi-granularity position assignment strategy to enhance the VLM's comprehension of visual information and improve the alignment between modalities.

\begin{figure*}[t]
  \centering 
    \includegraphics[width=\linewidth]{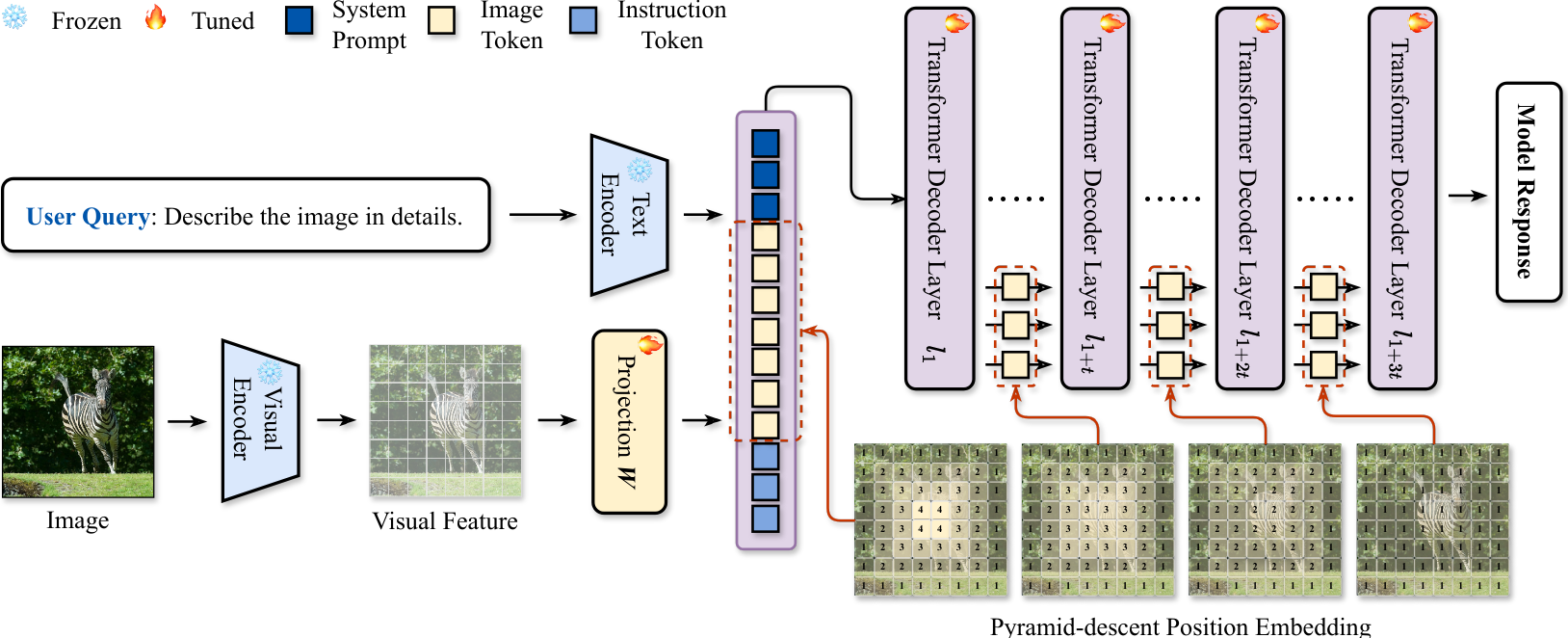}
\caption{\textbf{An overview of the proposed PyPE.} We first reorganize the visual tokens from their vanilla flattened 1D sequence form into the 2D format. Subsequently, we assign visual position indexes from the periphery to the center and expand the central receptive field incrementally across the layers with an interval of $t$.}
\label{fig:PyPE} 
\end{figure*}

\section{Approach}
\subsection{Preliminaries}
\label{sec:preliminaries}
\paragraph{RoPE (Rotary Position Embedding)} 
RoPE~\cite{su2024roformer} unifies both absolute and relative positional encodings, demonstrating a certain degree of extrapolation capability in LLMs and VLMs. Given the $m$-th query and $n$-th key vectors with a dimension $D$, denoted as $\mathbf{q}_m, \mathbf{k}_n \in \mathbb{R}^{|D|}$, RoPE multiplies a bias to the key or query vector in the complex vector space as follows:
\begin{equation}
    f_q(\mathbf{q}_m, m) = e^{im\Theta} \mathbf{q}_m, \quad
    f_k(\mathbf{k}_n, n) = e^{in\Theta} \mathbf{k}_n
    \label{eq:rope}
\end{equation}
where $\Theta = \mathrm{Diag}(\theta_1, \cdots, \theta_{|D|/2})$ is the rotary frequency matrix, where $\theta_d = b^{-2d/|D|}$ and the rotary base $b = 10000$. In real space, for $l = |D|/2$, the rotary matrix $e^{im\Theta}$ can be expressed as:
\begin{equation}
\setlength{\arraycolsep}{4pt}
    \begin{bmatrix}
        \cos m\theta_1 & -\sin m\theta_1 & \cdots & 0 & 0 \\
        \sin m\theta_1 & \cos m\theta_1 & \cdots & 0 & 0 \\
        \vdots & \vdots & \ddots & \vdots & \vdots \\
        0 & 0 & \cdots & \cos m\theta_l & -\sin m\theta_l \\
        0 & 0 & \cdots & \sin m\theta_l & \cos m\theta_l
    \end{bmatrix}
    \label{eq:real_space_rotary_matrix}
\end{equation}
The attention score using RoPE is calculated as follows:
\begin{equation}
\begin{aligned}
A_n &= \mathrm{Re}( f_q(\mathbf{q}_m, m), f_k(\mathbf{k}_n, n) ) \\
    &= \mathrm{Re}(\mathbf{q}^{\top}_m e^{i(m-n)\Theta} \mathbf{k}_n)
    \label{eq:rope_attn}
\end{aligned}
\end{equation}
where $\mathrm{Re}(\cdot)$ is the real part of a complex number and $e^{i(m-n)\Theta} = (e^{im\Theta})^{\top}e^{in\Theta}$. As the relative distance \(m-n\) increases, the attention score \(A_n\) correspondingly decreases due to long-term decay. This behavior aligns with the intuitive understanding that a pair of tokens separated by a significant relative distance should exhibit a weaker connection, and vice versa. However, a similar situation is observed in VLMs~\cite{xing2024mitigating}, which can lead to the model lacking attention to patches that are relatively far from the instruction token obtained through raster scanning.

\paragraph{All-One Position Encoding}
To further explore the impact of visual position encoding on the model's perception of visual elements, we propose All-One Position Encoding: directly setting the relative distance between all image tokens and instruction tokens to 1. By doing so, the relative distances from all image tokens to the instruction token become equal, thereby excluding the influence of relative position decay introduced by RoPE. As a result, all patches are treated equally.

As indicated in Table~\ref{tab: MME result}, All-One PE performs weaker than the baselines in perception but keeps competitive in coarse-grained perception tasks on different sizes of models. This suggests that even when assigning the same positional weight to all image tokens, the VLM still possesses certain perception capabilities and performs better than raster-scan and concentric in coarse-grained situations. This is more pronounced on LLaVA-1.5-13B because larger models have stronger sequence modeling and feature capturing capabilities, which correspondingly bridge the gap in fine-grained abilities between All-One PE and other methods.

\begin{table*}[t]
\centering
\setlength{\tabcolsep}{10pt} 
\resizebox{\linewidth}{!}{
\begin{tabular}{lccccccc}\toprule[1.2pt]
\multirow{2}{*}{\bf{Method}} & \multirow{2}{*}{\bf{Perception}} & \bf{Commonsense QA} & \multicolumn{4}{c}{\bf{Coarse-grained Perception Tasks}} &  \bf{Total} \\ \cmidrule(lr){4-7} 
& & \bf{(Reasoning)} & \bf{Existence} & \bf{Count} & \bf{Position} & \bf{Color} & \bf{Scores} \\ \midrule
 \multicolumn{2}{l}{\emph{{\textbf{TinyLLaVA-SigLIP-Phi-2}}}} \\
 \quad \textit{w/} Raster-scan &1488.30  &\bf{120.71} &185.00 &143.33 &133.33 &\bf{180.00} &762.37  \\
 \quad \textit{w/} Concentric &1465.25 &114.29 &185.00 &\bf{160.00} &131.67 &170.00 &760.96 \\
  \quad \textit{w/} All-One &1489.46 &117.14 &\bf{190.00} &158.33 &133.33 &175.00 &\bf{773.80} \\
\midrule
\rowcolor{aliceblue!80}   \quad \bf{\textit{w/} PyPE (Ours)} &\bf{1500.66} &115.00 &\bf{190.00} &150.00 &\bf{138.33} &\bf{180.00} &773.33 \\
\midrule
\multicolumn{2}{l}{\emph{{\textbf{LLaVA-1.5-7B}}}} \\
 \quad \textit{w/} Raster-scan & 1510.72  & \bf{135.71} & 190.00 & \bf{158.33} & 128.33 & 175.00 & 787.37  \\
 \quad \textit{w/} Concentric & 1485.67 & 120.71 & 190.00 & 153.33 & 133.33 & 170.00 & 767.37  \\
  \quad \textit{w/} All-One & 1474.13 & 131.43 & \bf{195.00} & 148.33 & 141.67 & 175.00 & 791.43 \\
  \midrule
\rowcolor{aliceblue!80}   \quad \bf{\textit{w/} PyPE (Ours)} & \bf{1542.19} & 130.00  & \bf{195.00} & 155.00 & \bf{146.67} & \bf{180.00} & \bf{806.67} \\
\midrule
\multicolumn{2}{l}{\emph{{\textbf{LLaVA-1.5-13B}}}} \\
  \quad \textit{w/} Raster-scan &1581.45   &\bf{132.14}  &190.00  &155.00  &135.00  &\bf{195.00}  &807.14   \\
 \quad \textit{w/} Concentric &1607.40   &128.57  &\bf{195.00}  &\bf{180.00}  &141.67  &185.00  &830.24  \\
  \quad \textit{w/} All-One &1608.12   &130.00  &\bf{195.00}  &170.00  &146.67  &190.00  &831.67 \\
  \midrule
\rowcolor{aliceblue!80}   \quad \bf{\textit{w/} PyPE (Ours)} &\bf{1629.41}   &130.71  &190.00  &\bf{180.00}  &\bf{153.33}  &180.00  &\bf{834.04} \\
 \bottomrule[1.2pt]
\end{tabular}
}
\caption{Performance evaluation on MME. \textit{Existence}, \textit{Count}, \textit{Position}, and \textit{Color} are coarse-grained subtasks of MME-Perception, while \textit{Commonsense QA} is a subtask of MME-Cognition. \textit{Total Scores} denotes the sum of the results from \textit{Commonsense QA} and \textit{Coarse-grained tasks}. The best results in each setting are in \bf{bold}.}
\label{tab: MME result}
\end{table*}

\subsection{Pyramid-descent Visual Position Encoding}
Though presenting competitive coarse-grained perception capabilities, All-One PE still falls short in fine-grained perception. Using identical position weights hampers the model's ability to differentiate the significance of image tokens, while the positional priors introduced by raster scanning conflict with general cognitive principles.

Similar challenges were also present in the early development of Vision Transformer (ViT)~\cite{dosovitskiy2020image}. Due to the columnar structure of ViT, which uses coarse image patches as input, it is difficult to apply it directly to pixel-level dense predictions such as object detection and segmentation. This difficulty arises because its output feature map is single-scale and low-resolution. To address these issues, \citet{wang2021pyramid} proposed the Pyramid Vision Transformer (PVT). They utilize fine-grained image patches as input to learn high-resolution representations and introduce a progressive shrinking pyramid to reduce the sequence length of the Transformer as the network deepens, significantly lowering the computational cost. Moreover, compared to CNNs, PVT consistently produces a global receptive field, ensuring a holistic perception of visual elements and benefiting its performance in detection and segmentation tasks.

\begin{algorithm}[t]
\caption{Pyramid-descent Visual Position Encoding}
\begin{algorithmic}[1] 

\item[\bf{INPUT:}] Height $H$, width $W$, descent interval $t$, current layer index $i$, current $\mathcal{P}_{max}$.
\item[\bf{OUTPUT:}] Pyramid-descent position assignment matrix $\mathcal{P}$, causal mask $\mathcal{M}$ and $\mathcal{P}_{max}$ for the next layer.

\IF{$i \mod t == 0 \ \ and \ \ \mathcal{P}_{max} > 1$}
    \STATE $\mathcal{P}_{max} \gets \mathcal{P}_{max} - 1$
\ENDIF

\STATE Initialize $\mathcal{P}$.

\FOR{$p \ \ in \ \ [1, \mathcal{P}_{max}]$}
    \STATE $\mathcal{P}[p : H - p, p : W - p] \gets p$
\ENDFOR

\STATE Generate $\mathcal{M}$ according to $\mathcal{P}$.

\end{algorithmic}
\end{algorithm}

\begin{table*}[t]
\centering
\setlength{\tabcolsep}{9pt} 
\resizebox{\textwidth}{!}{
\begin{tabular}{lccccccc}\toprule[1.2pt]
\bf{Method} & \bf{VQAv2} & \bf{OK-VQA$_\text{val}$} & \bf{GQA} &  \bf{VizWizQA} & \bf{TextVQA$_\text{val}$} & \bf{RealWorldQA} & \bf{ScienceQA$^\text{I}$} \\
\midrule
\multicolumn{2}{l}{\emph{\textbf{TinyLLaVA-SigLIP-Phi-2}}} \\
 \quad \textit{w/} Raster-scan &78.93  &56.71 &61.07 &34.30  &50.88 &53.99   &71.24 \\
 \quad \textit{w/} Concentric &79.08 &57.35 &61.15 &41.08 &50.77 &53.59 &70.45 \\
  \quad \textit{w/} All-One &78.89 &57.34 &61.33 &42.50 &50.94 &53.59 &70.55 \\
  \midrule
\rowcolor{aliceblue!80}   \quad \bf{\textit{w/} PyPE (Ours)} &\bf{79.22} &\bf{57.48} &\bf{61.65} &\bf{44.45} &\bf{51.31} &\bf{54.12} &\bf{71.59} \\
\midrule
\multicolumn{2}{l}{\emph{{\textbf{LLaVA-1.5-7B}}}} \\
  \quad \textit{w/} Raster-scan  &78.56  &54.32 &62.12 &50.34 &46.16 &54.80 &66.80  \\
 \quad \textit{w/} Concentric &79.02 &52.70 &62.28 &52.52 &45.84 &54.77 &68.72 \\
  \quad \textit{w/} All-One &79.02 &52.50 &62.00 &\bf{55.32} &45.98 & 54.77 &68.32 \\
  \midrule
\rowcolor{aliceblue!80}   \quad \bf{\textit{w/} PyPE (Ours)} &\bf{79.15} &\bf{54.96} &\bf{62.71} &53.11 &\bf{46.73} &\bf{55.42} &\bf{69.51} \\
  \midrule
  \multicolumn{2}{l}{\emph{{\textbf{LLaVA-1.5-13B}}}} \\
 \quad \textit{w/} Raster-scan &79.14  &\bf{57.38} &63.34 &53.75 &48.56 &55.95  &71.15 \\
 \quad \textit{w/} Concentric &79.90 &53.81 &63.26 &56.38 &48.07 & 55.42 &70.00 \\
  \quad \textit{w/} All-One &\bf{79.95} &51.40 &63.34 &56.37 &48.15 &54.64 &71.39 \\
  \midrule
\rowcolor{aliceblue!80}   \quad \bf{\textit{w/} PyPE (Ours)} &\bf{79.95} &55.66 &\bf{63.52} &\bf{58.06} &\bf{48.90} &\bf{56.86} &\bf{71.54} \\
 \bottomrule[1.2pt]
\end{tabular}
}
\caption{Performance evaluation on visual question answering. We utilize \textit{accuracy} as the evaluation metric. OK-VQA$_{\text{val}}$ and TextVQA$_{\text{val}}$ denote the validation set of OK-VQA and TextVQA, respectively. ScienceQA$^\text{I}$ denote the image subset of ScienceQA. The best results in each setting are in \bf{bold}.}
  \label{tab: VQA result}
\end{table*}

In light of this, we propose the \textbf{Pyramid-descent Visual Position Encoding (PyPE)}, a simple yet effective position assignment strategy for visual tokens in VLMs. As shown in Figure~\ref{fig:PyPE}, we first reorganize the visual tokens from their vanilla flattened 1D sequence form into the 2D format. Subsequently, we adopt a decay pattern for the corresponding position indexes of the image tokens that spread outward from the center following concentric PE~\cite{xing2024mitigating}. Given the maximum assignable position index $\mathcal{P}_{max}$, the position assignment matrix $\mathcal{P}$ is calculated as follows,
\begin{equation}
\begin{split}
&\mathcal{P}(i, j) = p, \ \ \forall p \in \left[1, \mathcal{P}_{max} \right], \\
s.t. \ \ \{(i, &j) \ | \ i \in [p,H-p) , \ j \in [p,W-p)\},
\end{split}
\end{equation}
where $H$ and $W$ represent the height and width of the input image, respectively. $\mathcal{P}_{max}$ is initialized to $\lfloor H/2 \rfloor$. This design maintains spatial continuity in the row and column dimensions. It reduces the average distance between significant image tokens and instruction tokens, facilitating cross-attention among the image tokens and cross-attention between the image tokens and instruction tokens. 

Subsequently, we propose a gradual expansion of the central receptive field to diminish the influence of anchor tokens and enhance the model's ability to perceive visual elements at varying levels of granularity. Specifically, we reduce $\mathcal{P}_{max}$ every $t$ layers, thereby controlling the granularity of perception through position encoding. When $\mathcal{P}_{max}$ is reduced to 1, the corresponding position encoding transforms into an All-One PE, which perceives more coarse-grained elements. To maintain causal attention, we adjust the attention mask $\mathcal{M}$ based on each assigned position matrix $\mathcal{P}$.

By introducing hierarchical position indices, PyPE facilitates multi-granularity perception of visual elements, allowing the model to dynamically adjust its focus to capture both broad contextual information and fine-grained details within visual data. This innovative approach not only aligns more closely with human cognitive processes but also enhances the model's overall performance in tasks that require both holistic and detailed perception of visual content.

\section{Experiment Setup}
\subsection{Benchmarks}
We evaluate PyPE on visual question answering and general multimodal benchmarks, including VQAv2~\cite{goyal2017making}, OK-VQA~\cite{marino2019ok}, GQA~\cite{hudson2019gqa}, VizWizQA~\cite{bigham2010vizwiz}, TextVQA~\cite{singh2019towards}, RealWorldQA~\cite{grok15}, ScienceQA~\cite{lu2022learn}, MME~\cite{yin2024survey}, MMBench~\cite{liu2025mmbench}, SEED-Bench~\cite{li2023seed}, POPE~\cite{li2023evaluating}, AI2D~\cite{kembhavi2016diagram}, MM-Vet~\cite{yu2023mm}, MMMU~\cite{yue2024mmmu}, MMT-Bench~\cite{ying2024mmt}, and MMStar~\cite{chen2024we}. Refer to Appendix~\ref{appendix:benchmarks} for more details.

\subsection{Implementation Details}
To demonstrate the generalizability of our proposed method across models with different parameter sizes, we conduct experiments using three model architectures with 3B, 7B, and 13B parameters. For 3B models, we follow TinyLLaVA~\cite{zhou2024tinyllava} to use SigLIP~\cite{zhai2023sigmoid} as the visual encoder and Phi-2~\cite{li2023textbooks} as the base LLM. For 7B and 13B models, we adopt pre-trained CLIP ViT-L/14 ($336^2$)~\cite{radford2021learning} as visual encoder and Vicuna v1.5~\cite{zheng2023judging} as the base LLM. Following~\citet{liu2024improved}, we pretrain the models on CC-558K dataset and finetune them on the mix-665K dataset. All experiments are conducted on 8 NVIDIA A100 and 8 NVIDIA H20 GPUs. See Appendix~\ref{appendix:hyperparameters} for more training and implementation details.

\begin{table*}[htbp]
\centering
\setlength{\tabcolsep}{6pt} 
\resizebox{\linewidth}{!}{
\begin{tabular}{lcccccccccccc}\toprule[1.2pt]
\multirow{2}{*}{\bf{Method}} & \multicolumn{3}{c}{\bf{POPE}} & \multicolumn{2}{c}{\bf{MMBench}}  & \multirow{2}{*}{\bf{SEED$^\text{I}$}}  & \multirow{2}{*}{\bf{AI2D}} & \multirow{2}{*}{\bf{MM-Vet}} & \multirow{2}{*}{\bf{MMMU}} & \multirow{2}{*}{\bf{MMT-Bench}} & \multirow{2}{*}{\bf{MMStar}}\\ \cmidrule(lr){2-4} \cmidrule(lr){5-6} & rand & pop & adv & en & cn & & & & & & \\
\midrule
\multicolumn{2}{l}{\emph{{\textbf{TinyLLaVA-SigLIP-Phi-2}}}} \\
 \quad \textit{w/} Raster-scan &88.50 &86.93 &85.60 &67.88 &\bf{45.07} &68.54 &59.75 &33.00 &\bf{33.80} &\bf{48.93}  &37.37 \\
 \quad \textit{w/} Concentric &88.63 &87.27 &85.67 &67.83 &43.22 &68.51 &60.98 &33.40 &33.60 &48.86  &38.44 \\
  \quad \textit{w/} All-One &88.53 &87.40 &86.00 &66.48 &43.11 &68.25 &61.20 &32.70 &\bf{33.80} &48.00  &38.06 \\
  \midrule
\rowcolor{aliceblue!80}   \quad \bf{\textit{w/} PyPE (Ours)} &\bf{89.07} &\bf{87.70} &\bf{85.73} &\bf{68.33} &43.95 &\bf{68.55} &\bf{61.53} &\bf{35.00} &33.70 &\bf{48.93} &\bf{38.89} \\
\midrule
\multicolumn{2}{l}{\emph{{\textbf{LLaVA-1.5-7B}}}} \\
  \quad \textit{w/} Raster-scan &\bf{88.33} &87.13 &85.63 &64.97 &57.90 &66.10 &55.25 &30.80 &31.00 &47.94  &35.25  \\
 \quad \textit{w/} Concentric &87.83 &86.40 &85.43 &65.41 &57.79 &66.31 &54.83 &29.70 &31.00 &49.02  &35.41 \\
  \quad \textit{w/} All-One &87.30 &86.57 &85.53 &65.47 &55.89 &66.41 &54.73 &29.90 &30.70 &48.99  &36.24 \\
  \midrule
\rowcolor{aliceblue!80}   \quad \bf{\textit{w/} PyPE (Ours)} &88.27 &\bf{87.43} &\bf{85.67} &\bf{66.65} &\bf{58.63} &\bf{67.01} &\bf{55.63} &\bf{31.10} &\bf{31.10} &\bf{49.70} &\bf{36.51}\\
  \midrule
  \multicolumn{2}{l}{\emph{{\textbf{LLaVA-1.5-13B}}}} \\
 \quad \textit{w/} Raster-scan &\bf{88.77} &\bf{87.70} &\bf{85.90} &67.74 &63.17 &67.65 &59.49 &\bf{37.30} &\bf{33.20} &49.82  &36.81 \\
 \quad \textit{w/} Concentric &87.90 &87.13 &85.80 &68.89 &62.67 &67.59 &58.55 &35.90 &32.70 &48.54  &37.33 \\
  \quad \textit{w/} All-One &87.93 &87.13 &85.77 &67.99 &63.06 &67.47 &58.84 &36.00 &32.90 &49.38  &37.32 \\
  \midrule
\rowcolor{aliceblue!80}   \quad \bf{\textit{w/} PyPE (Ours)} &88.03 &86.97 &85.47 &\bf{69.23} &\bf{63.45} & \bf{68.50} &\bf{59.59} &36.60 &\bf{33.20} &\bf{50.40}  &\bf{38.71} \\
 
 \bottomrule[1.2pt]
\end{tabular}
}
\caption{Evaluation on general multimodal benchmarks. We utilize \textit{accuracy} as the evaluation metric. SEED$^\text{I}$ denotes the image subset of SEED-Bench. The best results in each setting are in \bf{bold}.}
  \label{tab: MMB result}
\end{table*}

\begin{table}[t]
\centering
\setlength{\tabcolsep}{6pt} 
\resizebox{\linewidth}{!}{
\begin{tabular}{lcccc}\toprule[1.2pt]
\bf{Method} & \bf{MME$^\text{P}$}& \bf{OK-VQA$_\text{val}$} & \bf{TextVQA$_\text{val}$} & \bf{MMStar} \\
\midrule
\multicolumn{5}{c}{\emph{{\textbf{TinyLLaVA-SigLIP-Phi-2}}}} \\
  \midrule
 PyPE 1x &1479.53  &56.99 &50.13 &37.31   \\
\rowcolor{aliceblue!80}   \bf{PyPE 2x} &\bf{1500.66} &\bf{57.48} &\bf{51.31} &\bf{38.89}   \\
  PyPE 3x &1470.45 &57.29 &50.28 &38.32  \\
   PyPE 4x &1466.70 &55.89 &50.59 &37.23   \\
\midrule
\multicolumn{5}{c}{\emph{{\textbf{LLaVA-1.5-7B}}}} \\
\midrule
   PyPE 1x  &1507.19  &52.73 &\bf{46.77} &34.82\\
\rowcolor{aliceblue!80}   \bf{PyPE 2x} &\bf{1542.19} &\bf{54.96} &46.73 &\bf{36.51} \\
  PyPE 3x  &1503.95 &52.87 &45.99 &36.18 \\
   PyPE 4x &1497.18 &51.76 &46.20 &35.79  \\
  \midrule
  \multicolumn{5}{c}{\emph{{\textbf{LLaVA-1.5-13B}}}} \\
\midrule
  PyPE 1x &1608.01  &50.53 &48.60 &35.89   \\
\rowcolor{aliceblue!80}   \bf{PyPE 2x} &\bf{1629.41} &55.66 &\bf{48.90} &\bf{38.71}   \\
  PyPE 3x &1583.84 &54.90 &48.52 &36.55   \\
   PyPE 4x &1607.63 &\bf{57.42} &48.09 &37.07  \\
 \bottomrule[1.2pt]
\end{tabular}
}
\caption{Analysis of the descent interval $t$. PyPE $t$x denotes using PyPE with interval $t$. MME$^\text{P}$ denotes MME-Perception. }
  \label{tab: ablation_interval}
\end{table}

\section{Empirical Results and Analysis}
We evaluate the visual capabilities of the models trained with the PyPE through various visual question answering and general multimodal benchmarks. This novel position encoding demonstrates highly competitive performance at different scales. Our proposed method consistently delivers top-tier performance across most evaluation metrics, frequently surpassing other baselines.

\subsection{Results of Visual Question Answering Benchmarks}
To rigorously evaluate the capabilities of our models in general visual question answering tasks, we conduct comprehensive assessments across a diverse array of state-of-the-art benchmarks. The results presented in Tables~\ref{tab: MME result} and ~\ref{tab: VQA result} indicate that the PyPE series demonstrates exceptional performance across all benchmarks, with the three variants consistently achieving or surpassing baseline performance. In the MME benchmark, PyPE exhibits a superior understanding of visual content at various levels of granularity. It retains a coarse-grained perception capability comparable to that of All-One PE while outperforming both Raster-scan and Concentric PE in terms of fine-grained perception. On the RealWorldQA benchmark, which assesses real-world spatial comprehension, PyPE achieves scores of 54.12, 55.42, and 56.86 for the 3B, 7B, and 13B variants, respectively. These results exceed all baseline performances and reflect an enhanced understanding of physical environments. VizWizQA is a dataset comprising images captured by visually impaired individuals using mobile phones, accompanied by recorded spoken questions. The images in this dataset tend to exhibit relatively low clarity, with subjects occupying a significant portion of the frame. Consequently, as shown in Table~\ref{tab: VQA result}, All-One PE demonstrates competitive performance on this dataset, while our proposed PyPE exhibits superior zero-shot performance on both VizWizQA and ScienceQA. This improvement can be attributed to the flexible receptive field enabled by PyPE.

\begin{figure*}[t]
  \centering 
    \includegraphics[width=\linewidth]{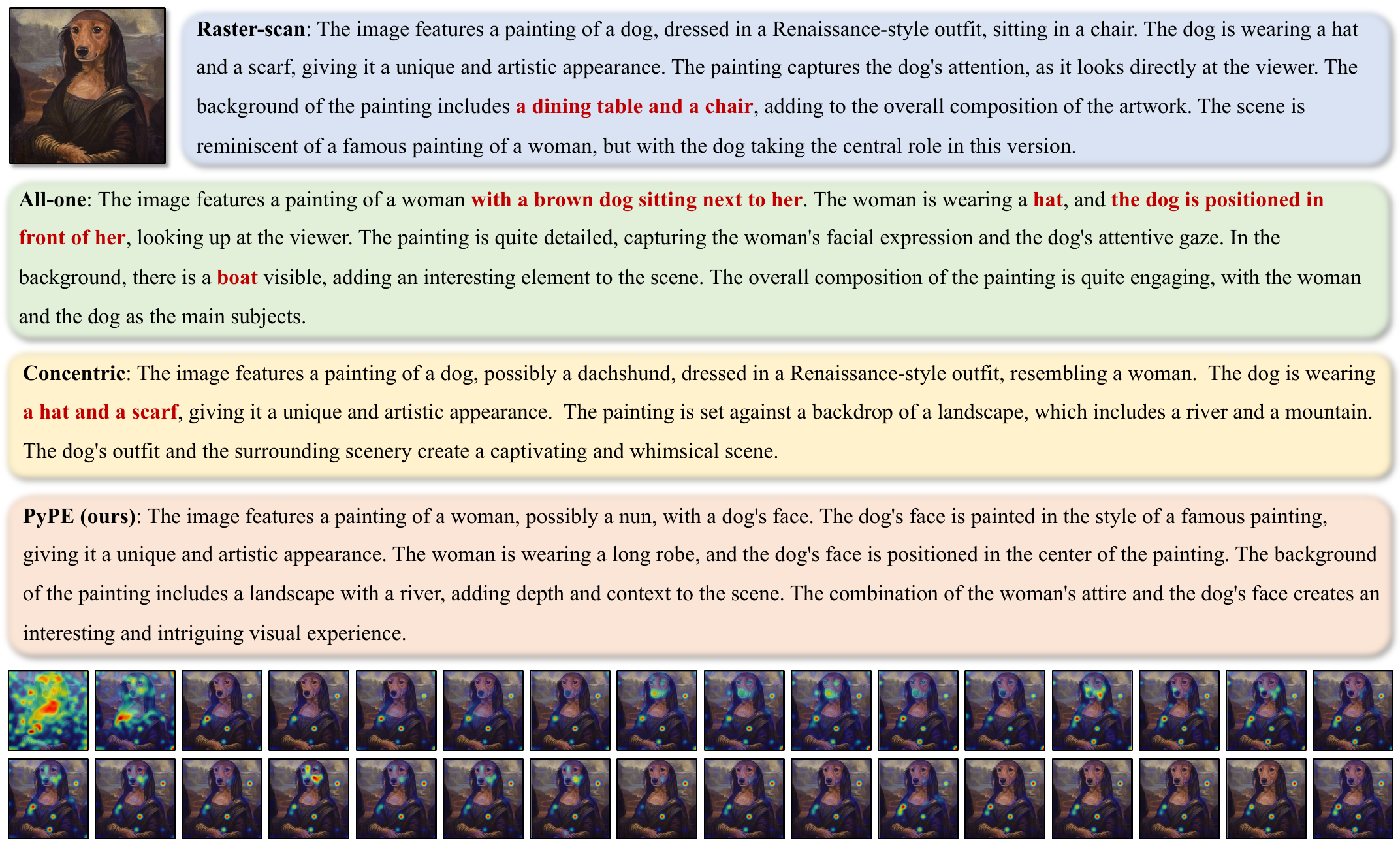}
\caption{Illustration of the multi-granularity perception capability of PyPE with a sample from LLaVA-Bench. The case study is based on LLaVA-1.5-7B and the query is \textit{"Describe this photo in detail"}. The misunderstandings and hallucinations of visual contents are highlighted in \textbf{\textcolor{BrickRed}{red}}. We also provide a corresponding layer-wise attention visualization of PyPE, with the heatmap arranged from the upper left to the lower right, indicating layers 1 to 32.}
\label{fig:case}
\end{figure*}

\subsection{Results of General Multimodal Benchmarks}
As illustrated in Table~\ref{tab: MMB result}, the PyPE series demonstrates exceptional performance on mainstream general multimodal benchmarks. In the MMStar benchmark, which is designed to assess genuine multimodal capabilities using visually indispensable samples, PyPE outperforms all baseline models. On MM-Vet, which evaluates the integration of core vision-language capabilities across 16 complex multimodal tasks, the 3B model of PyPE achieves an impressive score of 35.00, significantly surpassing the scores of 33.00 and 33.40 obtained by Raster-scan and Concentric PE, respectively. In the MMT-Bench evaluation, which assesses advanced reasoning and instruction-following across 32 core meta-tasks and 162 subtasks in multimodal understanding, PyPE markedly exceeds baseline performance, demonstrating its ability to apply expert knowledge and execute deliberate visual recognition, localization, reasoning, and planning. On MMBench, which evaluates fine-grained abilities across 20 dimensions, PyPE exhibits strong performance, matching or leading the state-of-the-art. Additionally, we test the methods on AI2D, a benchmark focusing on multiple-choice questions related to scientific diagrams containing text. The results indicate that PyPE achieves state-of-the-art performance and demonstrates a strong comprehension of textual content within images.

\subsection{Analysis of the Descent Interval}
As shown in Table~\ref{tab: ablation_interval}, we evaluate the performance of different models using PyPE with varying descent intervals on VQA and general multimodal benchmarks. Across all models, a moderate descent interval PyPE 2x generally provides the best or near-best performance, which strikes a balance between the model's ability to handle perception (MME), external knowledge integration (OK-VQA), text comprehension (TextVQA), and vision-critical tasks (MMStar). While the 2x interval is generally optimal, there are exceptions, such as the LLaVA-1.5-13B model performing best on OK-VQA with a 4x interval. This indicates that larger models might benefit from longer intervals for specific tasks. 



\subsection{Qualitative Results on LLaVA-Bench}
\label{sec:case}
Figure~\ref{fig:case} demonstrates a case study on how, given identical prompts and images, other baselines misperceive or inadequately process visual information, resulting in the generation of hallucinatory content. For instance, in the displayed example, the baseline methods exhibit object hallucinations, identifying nonexistent items such as "\textit{dining table}", "\textit{hat}", "\textit{scarf}", and "\textit{boat}". In contrast, the implementation of PyPE notably mitigates these hallucination issues while simultaneously maintaining the coherence and informativeness of the output text. This can be attributed to the multi-scale visual modeling capability afforded by the dynamic local receptive fields of PyPE, in conjunction with the stable global receptive fields. Furthermore, the visualization results of layer-wise attention indicate that our proposed method effectively alleviates the phenomenon of "aggregation pattern", thereby creating a synergistic effect with the former. 
Refer to Appendix~\ref{appendix:anchor} for a more in-depth analysis of anchor tokens and Appendix~\ref{appendix:more_cases} for more case studies.

\section{Conclusion}

In this work, we conduct an in-depth analysis of how visual position encoding affects visual perception in VLMs, particularly from the aspect of long-term decay and the "aggregation pattern". We find that conventional visual position encoding methods are constrained by the "aggregation pattern" derived from LLMs and lack multi-scale perceptual capabilities. To address these limitations, we introduce Pyramid-descent Visual Position Encoding (PyPE), a novel approach designed to enhance the perception of visual tokens within VLMs. Extensive experiments across multiple benchmarks and VLM families demonstrate the efficacy of PyPE in addressing these challenges and ensuring a thorough understanding of visual content.

\section*{Limitations}
Although PyPE demonstrates exceptional performance in enhancing the overall capabilities of Vision-language Models (VLMs), it is currently limited to single-frame images and has not yet been extended to video and other modalities. Future research will focus on effectively integrating the temporal dimension for unified position encoding and extending PyPE to a broader range of VLMs.

\bibliography{custom}

\appendix
\section{Benchmarks}
\label{appendix:benchmarks}

\begin{figure*}[!ht]
  \centering 
\subfloat[Raster-scan PE.]{
    \includegraphics[width=0.45\linewidth]{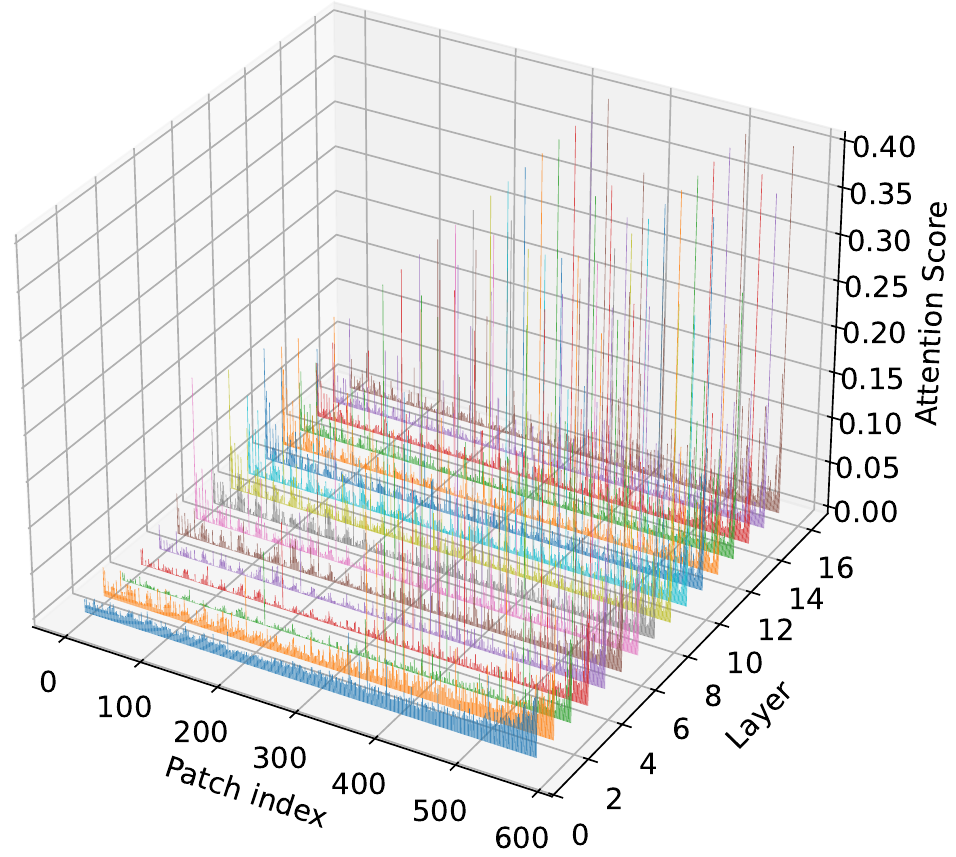}
  }
  \hfill
\subfloat[All-One PE.]{
    \includegraphics[width=0.45\linewidth]{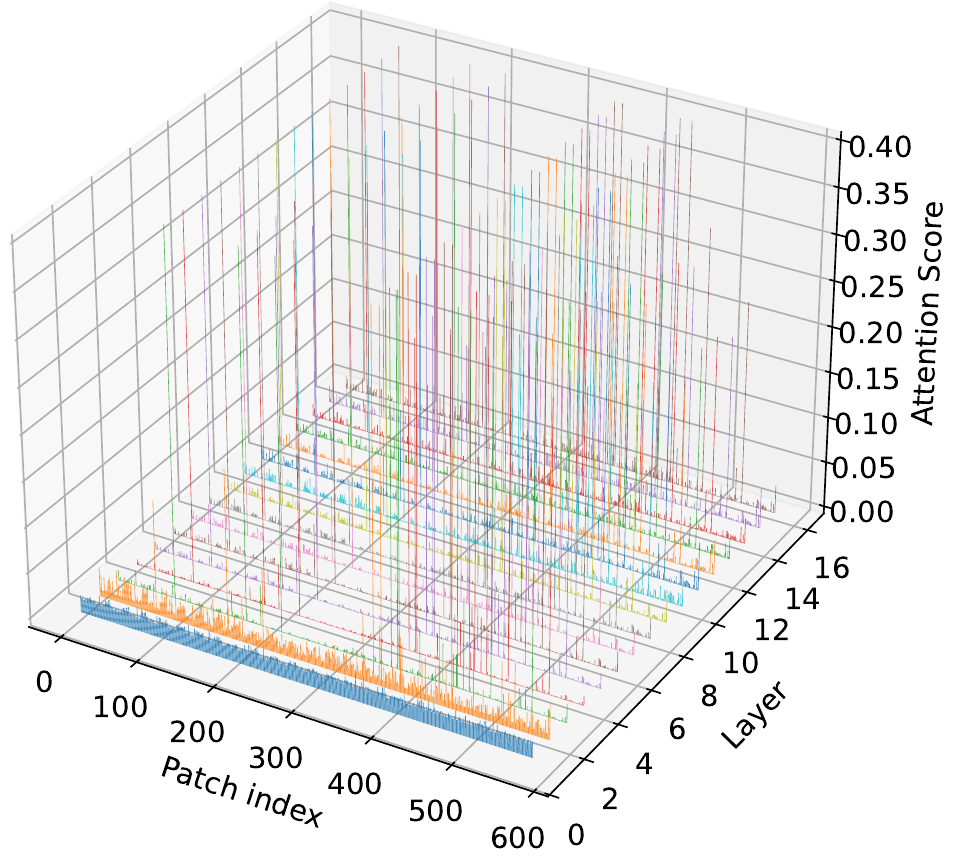}
  }
  \vspace{1em}
\subfloat[Concentric PE.]{
    \includegraphics[width=0.45\linewidth]{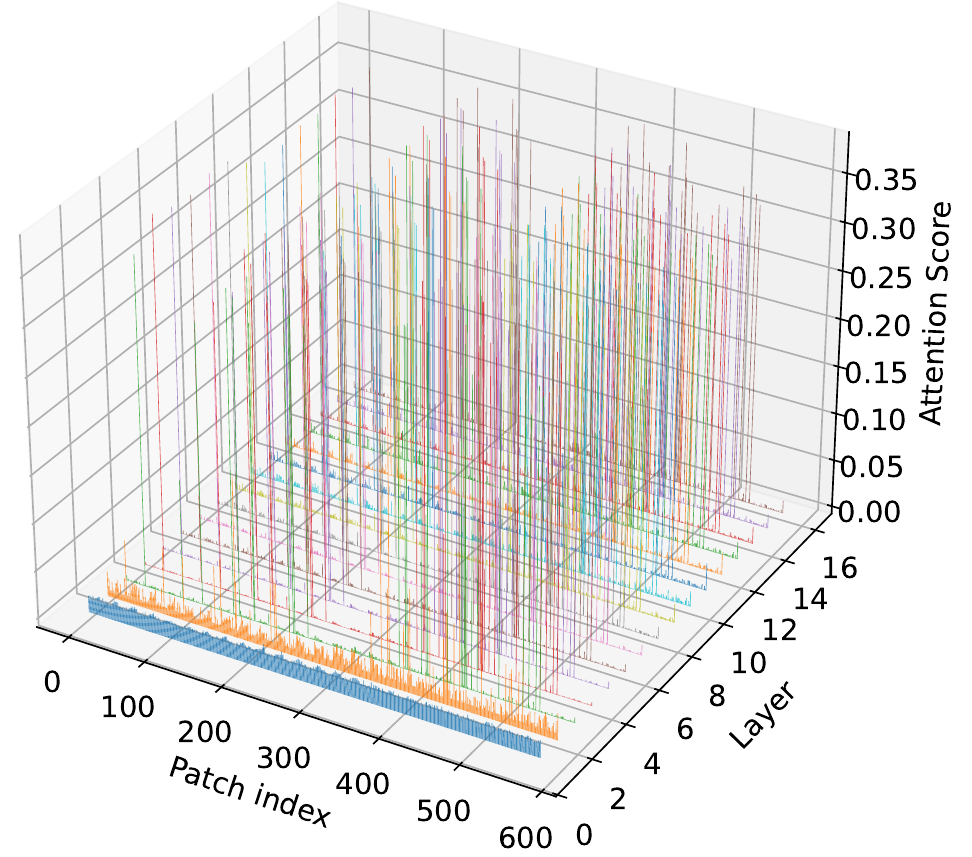}
  }
  \hfill
\subfloat[PyPE.]{
    \includegraphics[width=0.45\linewidth]{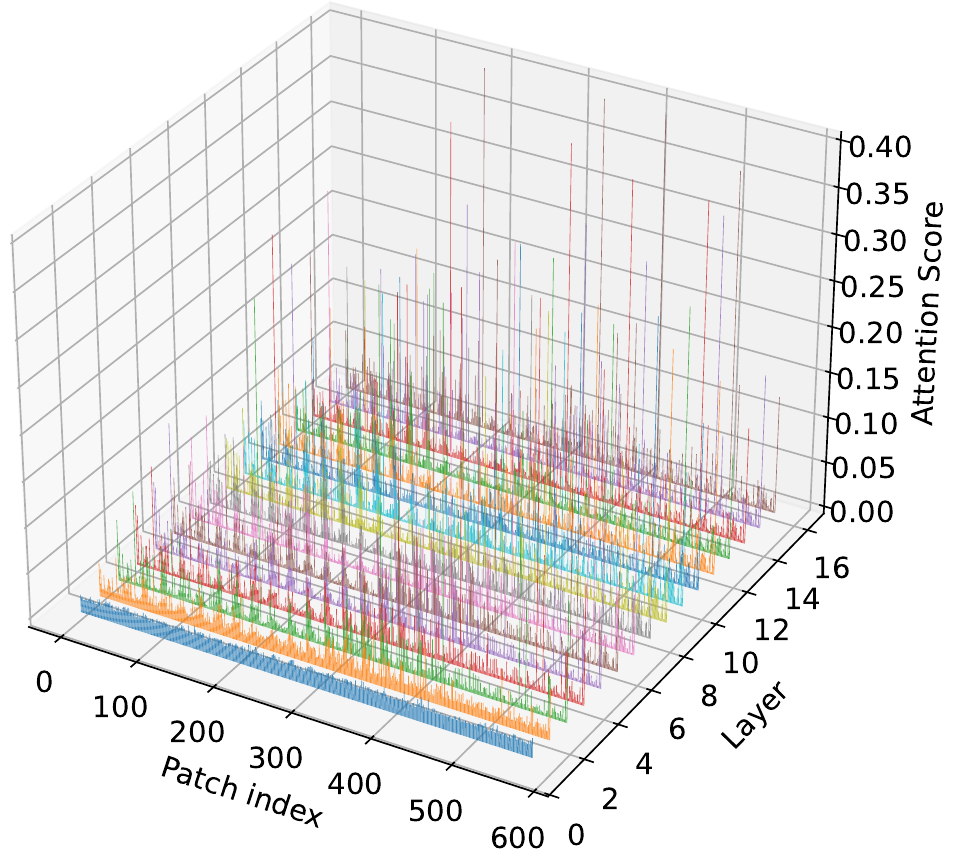}
  }
\caption{Visualization of anchor tokens in baselines and PyPE.} 
\label{fig:PE_3d} 
\end{figure*}

\paragraph{Visual Question Answering}
The VQAv2 dataset is currently the largest available dataset for visual question answering. OK-VQA includes questions that necessitate external knowledge beyond the multimodal inputs provided. GQA is specifically designed to assess the reasoning capabilities of the model. VizWizQA is composed of question-answer pairs derived from visually impaired users. TextVQA places a greater emphasis on evaluating the model's ability to comprehend text within natural scenes. RealWorldQA is a benchmark specifically designed to evaluate the spatial understanding capabilities of multimodal AI models in real-world contexts. ScienceQA comprises multimodal multiple-choice questions across a diverse range of science topics. These datasets are strategically selected to comprehensively evaluate our method's capacity to understand and reason across diverse visual contexts and knowledge domains.

\paragraph{General Multimodal Benchmarks}
MME measures both perception and cognition abilities on a total of 14 subtasks. MMBench comprehensively evaluates a model's multimodal capabilities in both Chinese and English contexts. SEED-Bench focuses on assessing generative comprehension in Vision-language Models. POPE evaluates the extent of multimodal hallucinations present in a model. AI2D assesses a model's ability to interpret scientific diagram inputs. MM-Vet evaluates the multimodal conversational abilities of a model using GPT-4 as a benchmark. MMMU is designed to assess multimodal models on extensive multi-disciplinary tasks that require college-level subject knowledge and deliberate reasoning. MMT-Bench is a comprehensive benchmark developed to evaluate VLMs across a wide range of multimodal tasks that necessitate expert knowledge and deliberate visual recognition, localization, reasoning, and planning. MMstar is a premier, vision-critical multimodal benchmark comprising 1,500 challenge samples meticulously curated by human experts.

\begin{table}[t]
\centering
\setlength{\tabcolsep}{16pt} 
\resizebox{\linewidth}{!}{
\begin{tabular}{l| c c}
\toprule[1.2pt]
 Hyperparameter & Pretrain & Finetune \\
\midrule
batch size & 256 & 128 \\
 lr & 1e-3 & 2e-5 \\
 lr schedule & \multicolumn{2}{c}{cosine decay} \\
 lr warmup ratio & \multicolumn{2}{c}{0.03} \\
 weight decay & \multicolumn{2}{c}{0} \\
 epoch & \multicolumn{2}{c}{1} \\
 optimizer & \multicolumn{2}{c}{AdamW} \\
 DeepSpeed stage & 2 & 3 \\
\bottomrule[1.2pt]
\end{tabular}
}
\caption{
\textbf{Hyperparameters} of TinyLLaVA-SigLIP-Phi-2 and LLaVA-1.5-7B/13B.}
\label{tab:hyperparameter}
\end{table}

\begin{table*}[t]
\centering
\setlength{\tabcolsep}{14pt} 
\resizebox{\linewidth}{!}{
\begin{tabular}{lcccccccc}\toprule[1.2pt]
\multirow{2}{*}{\bf{Method}} & \multicolumn{3}{c}{\bf{RefCOCO}} & \multicolumn{3}{c}{\bf{RefCOCO+}}  & \multicolumn{2}{c}{\bf{RefCOCOg}}\\ \cmidrule(lr){2-4} \cmidrule(lr){5-7} \cmidrule(lr){8-9} & val & test-A & test-B & val & test-A & test-B & val & test \\
\midrule
\multicolumn{2}{l}{\emph{{\textbf{TinyLLaVA-SigLIP-Phi-2}}}} \\
 \quad \textit{w/} Raster-scan &\bf{31.85} &15.77 &\underline{44.29} & \underline{31.28} & 18.65 & \underline{40.83} & 56.86 & 56.98 \\
 \quad \textit{w/} Concentric &30.89 &\underline{15.97} &42.22 &31.16 &\underline{19.31} &38.77 &\underline{59.45} &\underline{58.74}   \\
  \quad \textit{w/} All-One &29.13 &15.24 &39.11 &30.64 &18.38 &38.34 &54.81 &55.28   \\
  \midrule
\rowcolor{aliceblue!80}   \quad \bf{\textit{w/} PyPE (Ours)} &\underline{31.33} &\bf{16.02} &\bf{45.13} &\bf{31.86} &\bf{19.40} &\bf{42.25} &\bf{59.72} &\bf{59.79}   \\
\midrule
\multicolumn{2}{l}{\emph{{\textbf{LLaVA-1.5-7B}}}} \\
  \quad \textit{w/} Raster-scan &34.19 &\bf{18.07} &46.89 &\bf{34.30} &\bf{21.63} &\bf{43.53} &61.21 &59.40    \\
 \quad \textit{w/} Concentric &32.23 &16.51 &42.49 &32.66 &20.00 &40.41 &59.72 &58.47   \\
  \quad \textit{w/} All-One &32.99 &16.46 &41.26 &33.28 &20.73 &39.83 &63.07 &61.90   \\
  \midrule
\rowcolor{aliceblue!80}   \quad \bf{\textit{w/} PyPE (Ours)} &\bf{35.16} &16.51 &\bf{48.04} &34.17 &21.22 &41.46 &\bf{64.62} &\bf{64.13}  \\
  \midrule
  \multicolumn{2}{l}{\emph{{\textbf{LLaVA-1.5-13B}}}} \\
 \quad \textit{w/} Raster-scan &36.86 &19.29 &50.01 &36.12 &22.37 &43.59 &\bf{63.66} &60.96   \\
 \quad \textit{w/} Concentric &35.87 &18.54 &48.17 &36.07 &21.94 &42.65 &61.66 &61.07  \\
  \quad \textit{w/} All-One &36.84 &19.06 &49.16 &37.10 &22.71 &41.72 &61.58 &59.75   \\
  \midrule
\rowcolor{aliceblue!80}   \quad \bf{\textit{w/} PyPE (Ours)} &\bf{37.81} &\bf{21.82} &\bf{51.88} &\bf{37.14} &\bf{25.74} &\bf{44.73} &63.16 &\bf{62.59}  \\
 
 \bottomrule[1.2pt]
\end{tabular}
}
\caption{Performance comparison on referring expression comprehension tasks. We use CIDEr~\cite{vedantam2015cider} to evaluate the quality of the descriptions. The highest results in each setting are indicated in \textbf{bold}, while the second-best results are \underline{underlined}.}
  \label{tab: refer_result}
\end{table*}

\section{Hyperparameters and More Implementation Details}
\label{appendix:hyperparameters}
We show the training hyperparameters for both first-stage vision-language alignment pretraining and the second-stage visual instruction tuning in Table~\ref{tab:hyperparameter}. We use LMMs-Eval~\cite{zhang2024lmms} to conduct experiments on VQA and general multimodal benchmarks.

\section{Visualization of Anchor Tokens}
\label{appendix:anchor}
To further analyze the aggregating attention pattern, we visualize the attention score of each patch in the first 16 layers. As illustrated in Figure~\ref{fig:PE_3d}, both the All-One PE and the Concentric PE exhibit a relatively uniform distribution of attention in the initial two layers. However, a significant phenomenon of attention aggregation emerges in the subsequent layers, where non-anchor patches demonstrate a suppression of attention, particularly pronounced in Concentric PE. Though Raster-scan PE shows slight improvement, the attention in each layer tends to be preferentially allocated to patches that are closer to the instruction token, resulting in a discontinuous and fragmented attention pattern. This indicates a limitation of the Raster-scan PE in effectively modeling patches with similar semantics. In contrast, PyPE not only reduces the number of anchor tokens but also yields significantly lower attention scores for these tokens compared to the baselines, thereby facilitating the model's exploration of image details more effectively. Furthermore, in each layer, the attention distribution of the PyPE is more continuous, highlighting the superiority of our proposed method in modeling semantically similar information.

\section{Performance on Referring Expression Comprehension}
\label{appendix:refcoco}
In the context of the visual localization task, we evaluate PyPE using the RefCOCO, RefCOCO+, and RefCOCOg datasets~\cite{kazemzadeh2014referitgame,mao2016generation}. The results, presented in Table~\ref{tab: refer_result}, indicate that PyPE achieves top-tier performance among baselines. Its superior structural design enables PyPE to effectively perceive intricate details within images, resulting in significant improvements over baseline models. The performance of PyPE underscores its potential to advance the field of visual localization and its applicability in real-world scenarios that require precise visual understanding.

\section{More Case Studies}
\label{appendix:more_cases}
\begin{figure*}[t]
    \centering
        \begin{minipage}[b]{.65\textwidth}
        \centering
            \includegraphics[width=\textwidth]{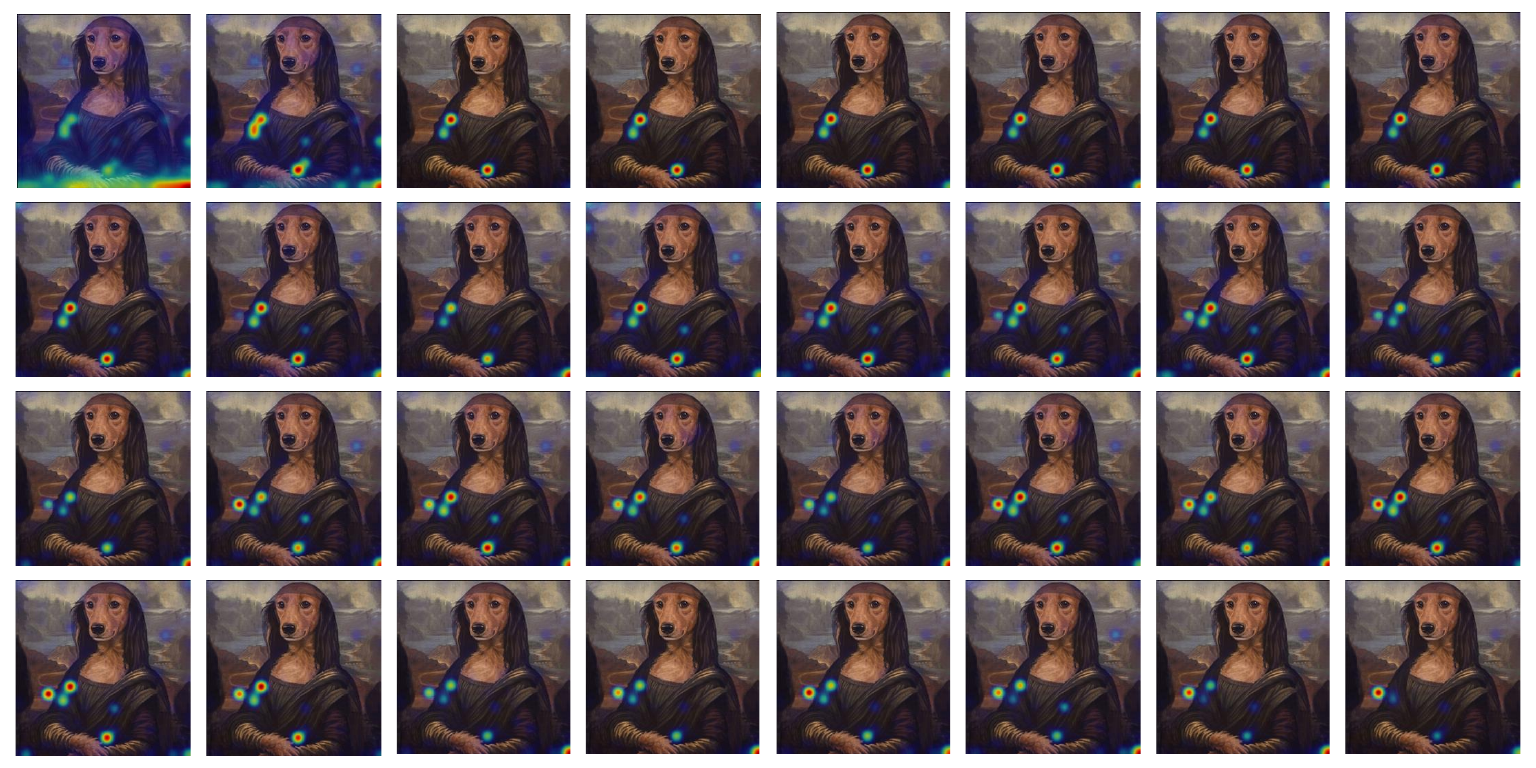}
            \subcaption{Raster-scan}
        \end{minipage}

        \begin{minipage}[b]{.65\textwidth}
        \centering
            \includegraphics[width=\textwidth]{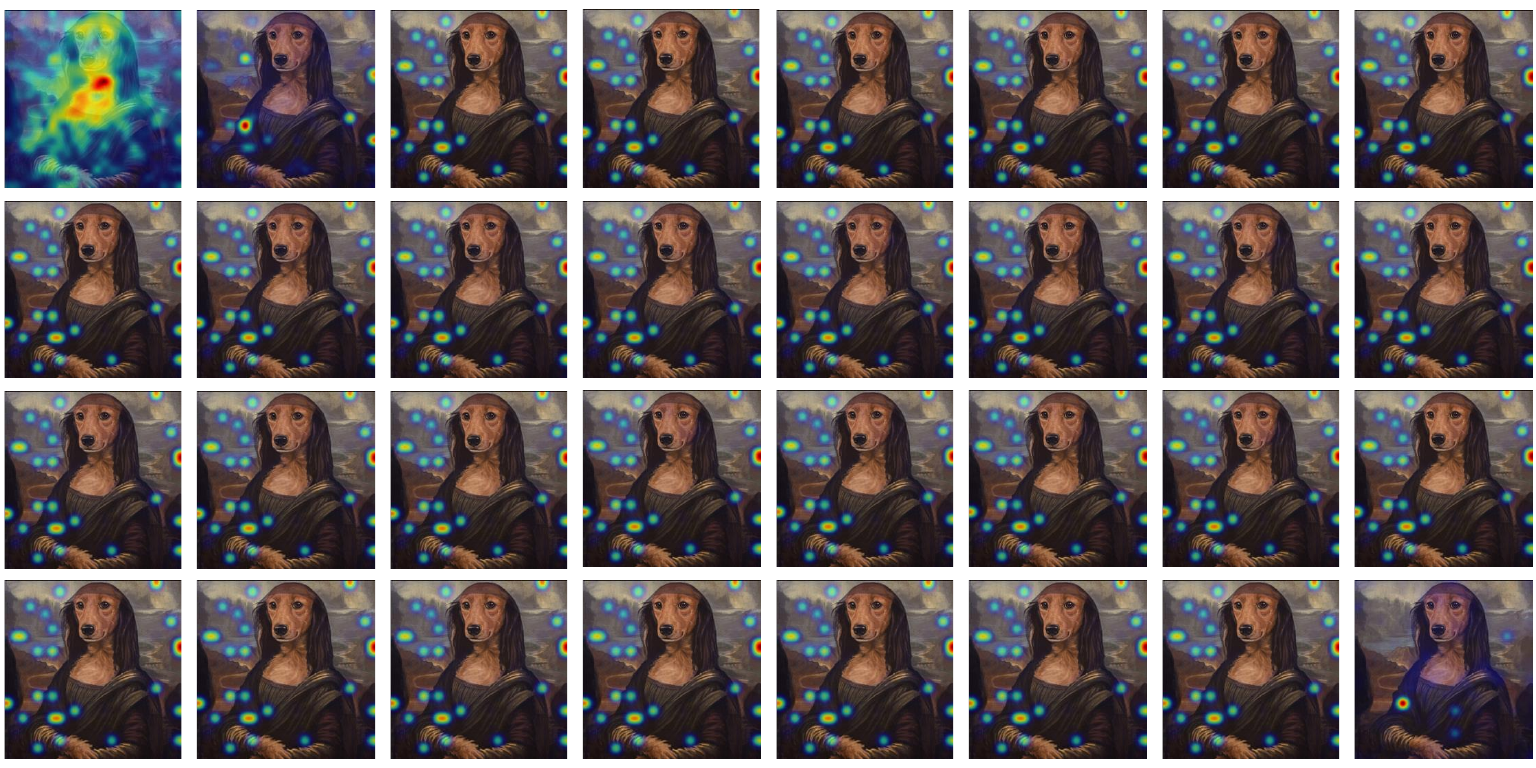}
            \subcaption{Concentric}
        \end{minipage}

        \begin{minipage}[b]{.65\textwidth}
        \centering
            \includegraphics[width=\textwidth]{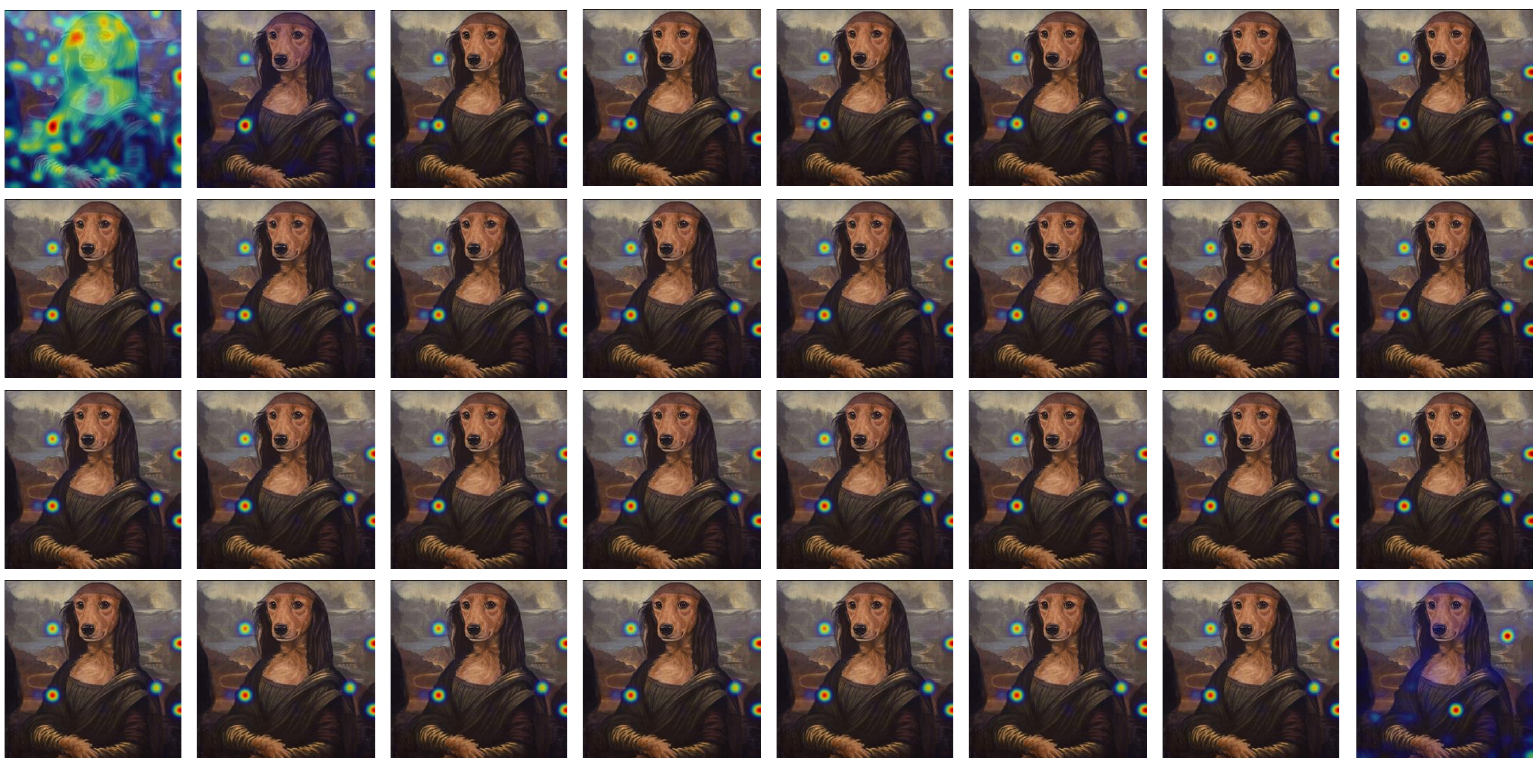}
            \subcaption{All-One}
        \end{minipage}

        \begin{minipage}[b]{.65\textwidth}
        \centering
            \includegraphics[width=\textwidth]{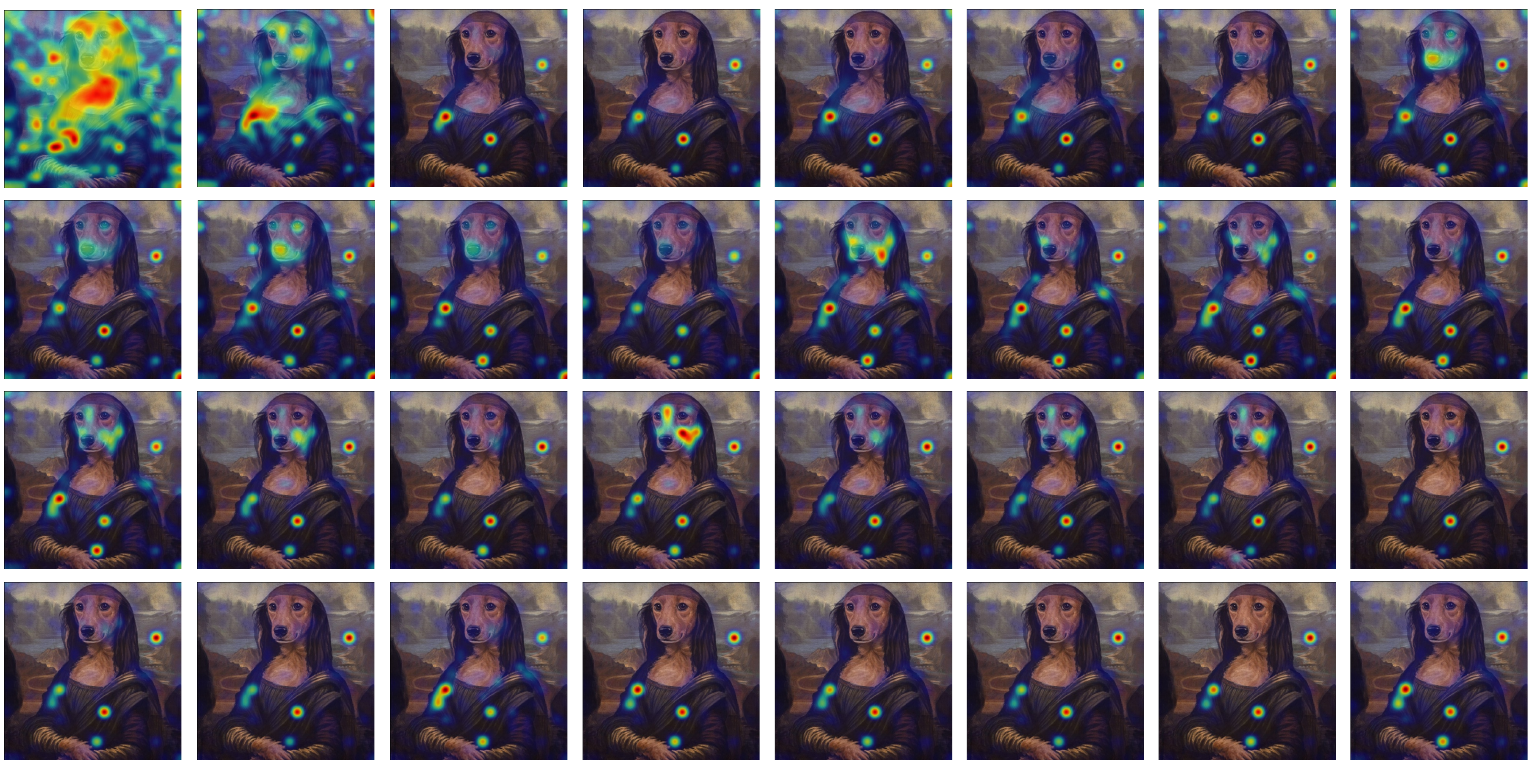}
            \subcaption{PyPE}
        \end{minipage}
    \caption{Layer-wise attention visualization (left to right, up to down) of the example from Figure~\ref{fig:case}.}
    \label{fig:more_vis_0}
\end{figure*}

\begin{figure*}[t]
    \centering
        \begin{minipage}[b]{\textwidth}
        \centering
            \includegraphics[width=\textwidth]{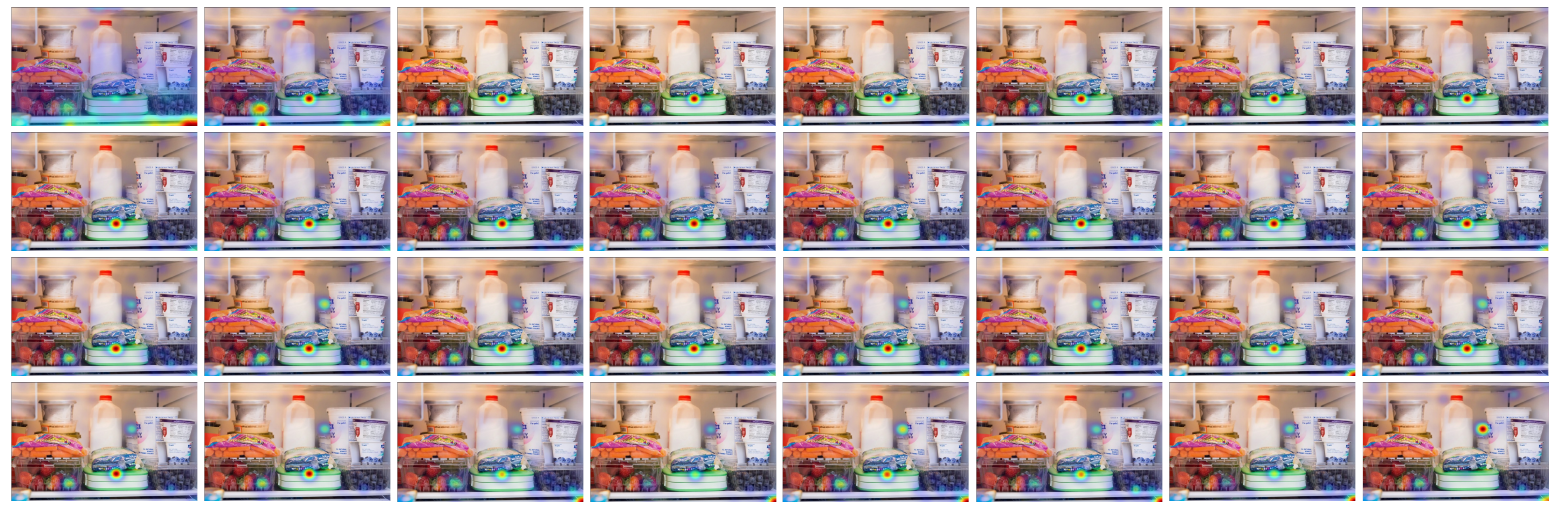}
            \subcaption{Raster-scan}
        \end{minipage}

        \begin{minipage}[b]{\textwidth}
        \centering
            \includegraphics[width=\textwidth]{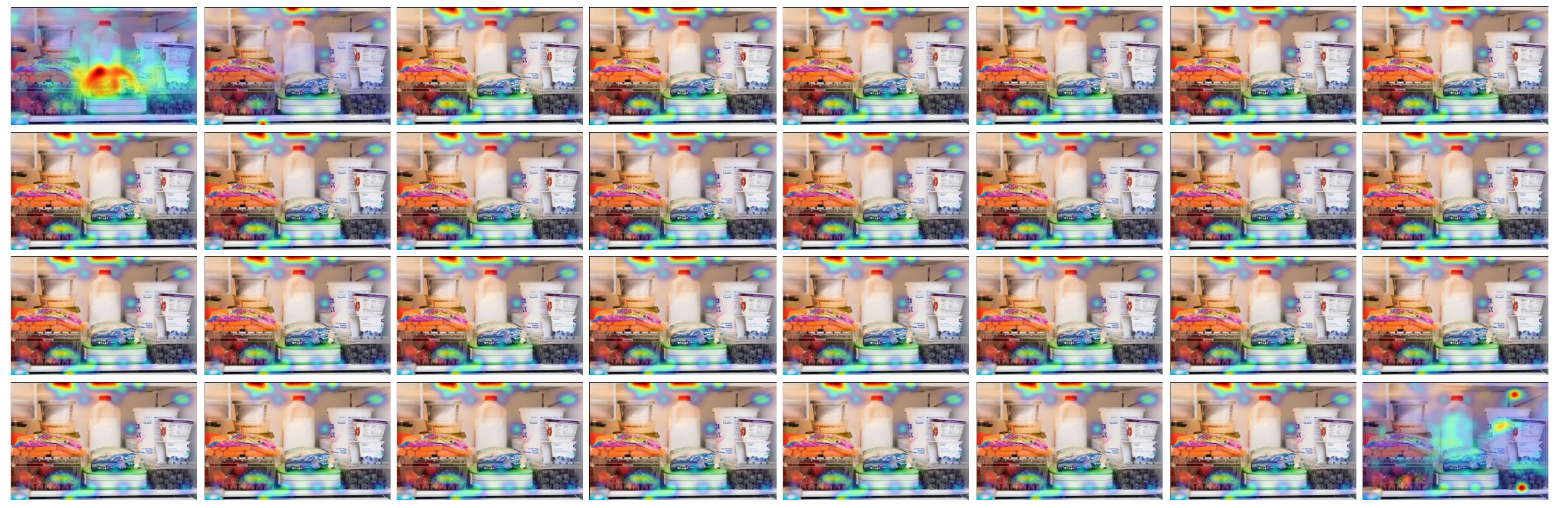}
            \subcaption{Concentric}
        \end{minipage}

        \begin{minipage}[b]{\textwidth}
        \centering
            \includegraphics[width=\textwidth]{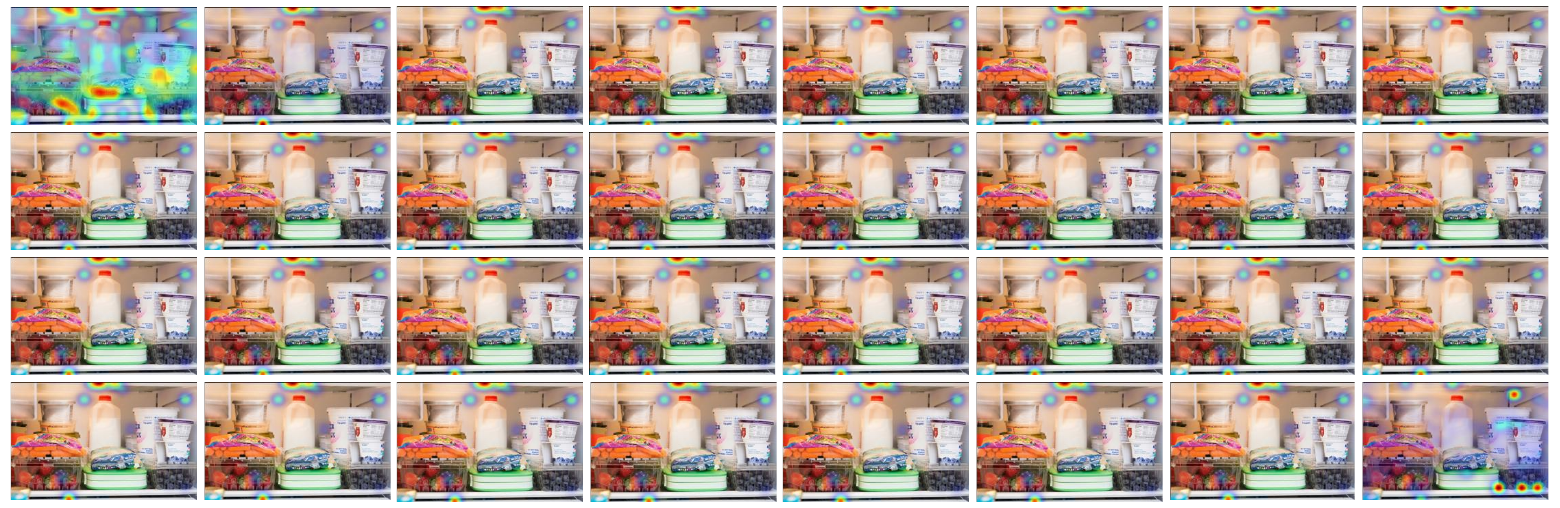}
            \subcaption{All-One}
        \end{minipage}

        \begin{minipage}[b]{\textwidth}
        \centering
            \includegraphics[width=\textwidth]{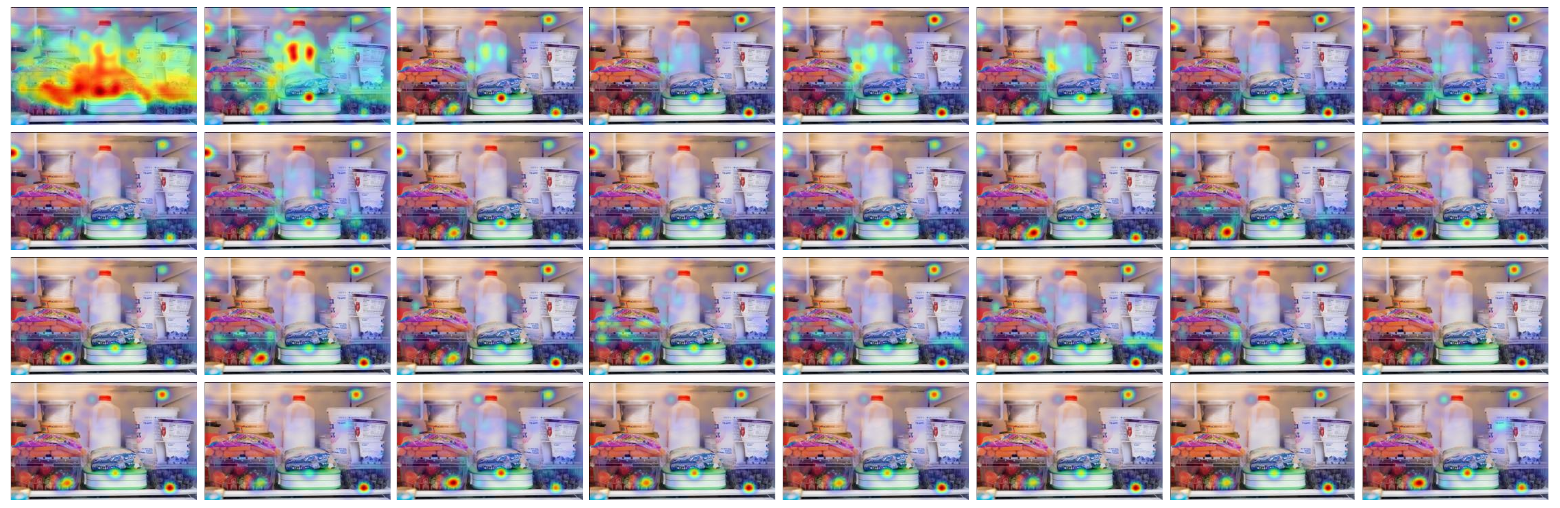}
            \subcaption{PyPE}
        \end{minipage}
    \caption{Layer-wise attention visualization (left to right, up to down) of the first example from Table~\ref{tab:more_cases}.}
    \label{fig:more_vis_1}
\end{figure*}

\begin{figure*}[t]
    \centering
        \begin{minipage}[b]{.65\textwidth}
        \centering
            \includegraphics[width=\textwidth]{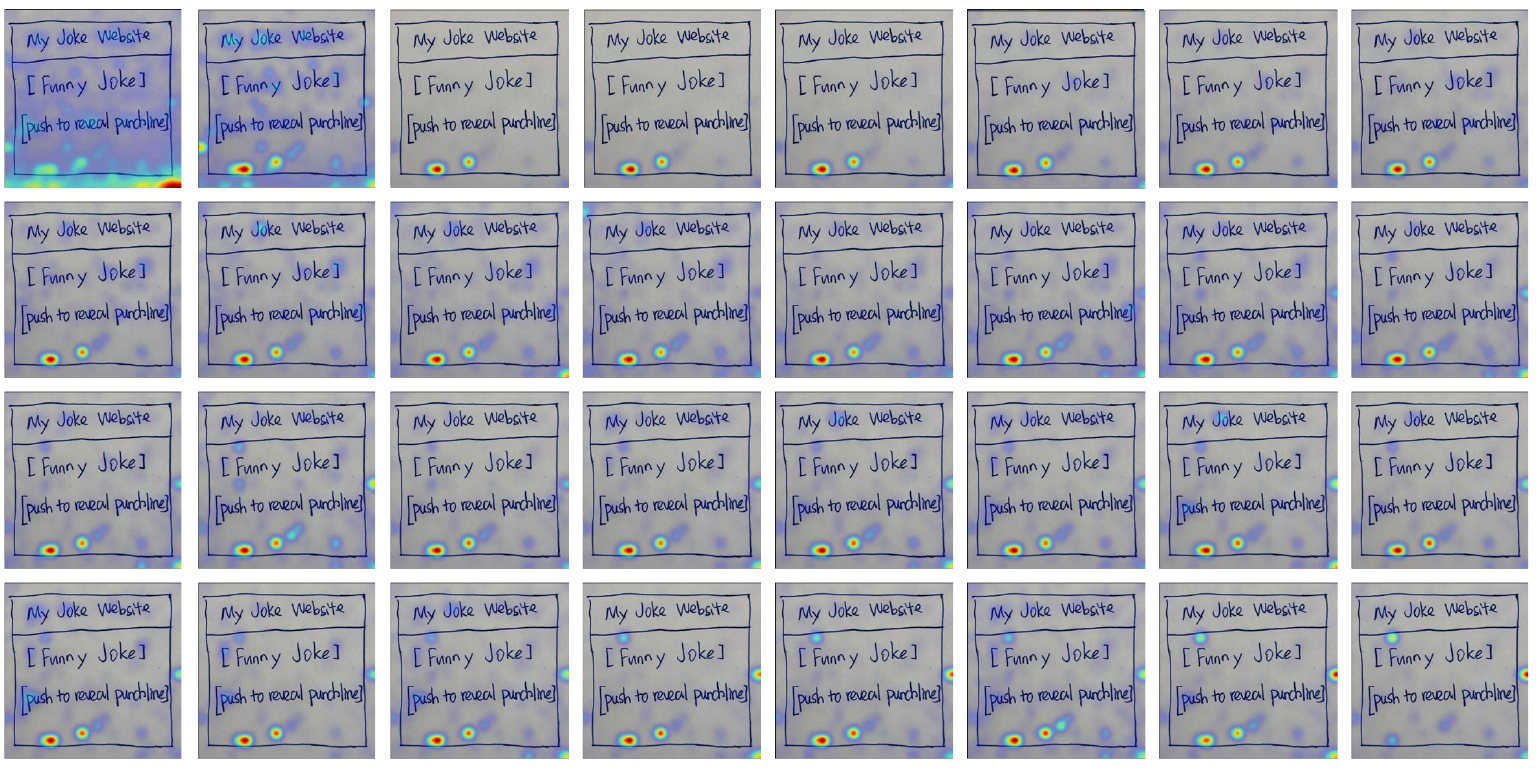}
            \subcaption{Raster-scan}
        \end{minipage}

        \begin{minipage}[b]{.65\textwidth}
        \centering
            \includegraphics[width=\textwidth]{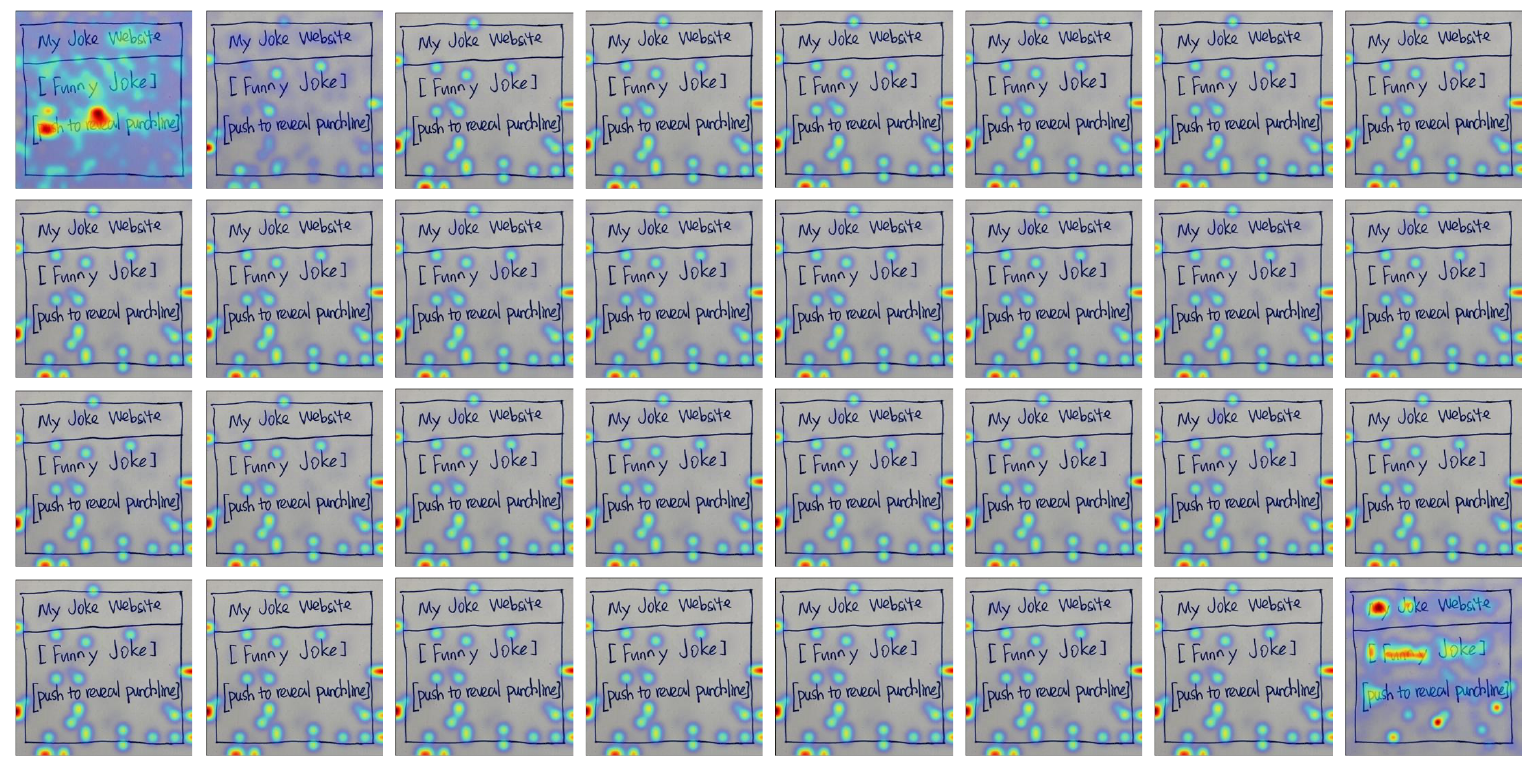}
            \subcaption{Concentric}
        \end{minipage}

        \begin{minipage}[b]{.65\textwidth}
        \centering
            \includegraphics[width=\textwidth]{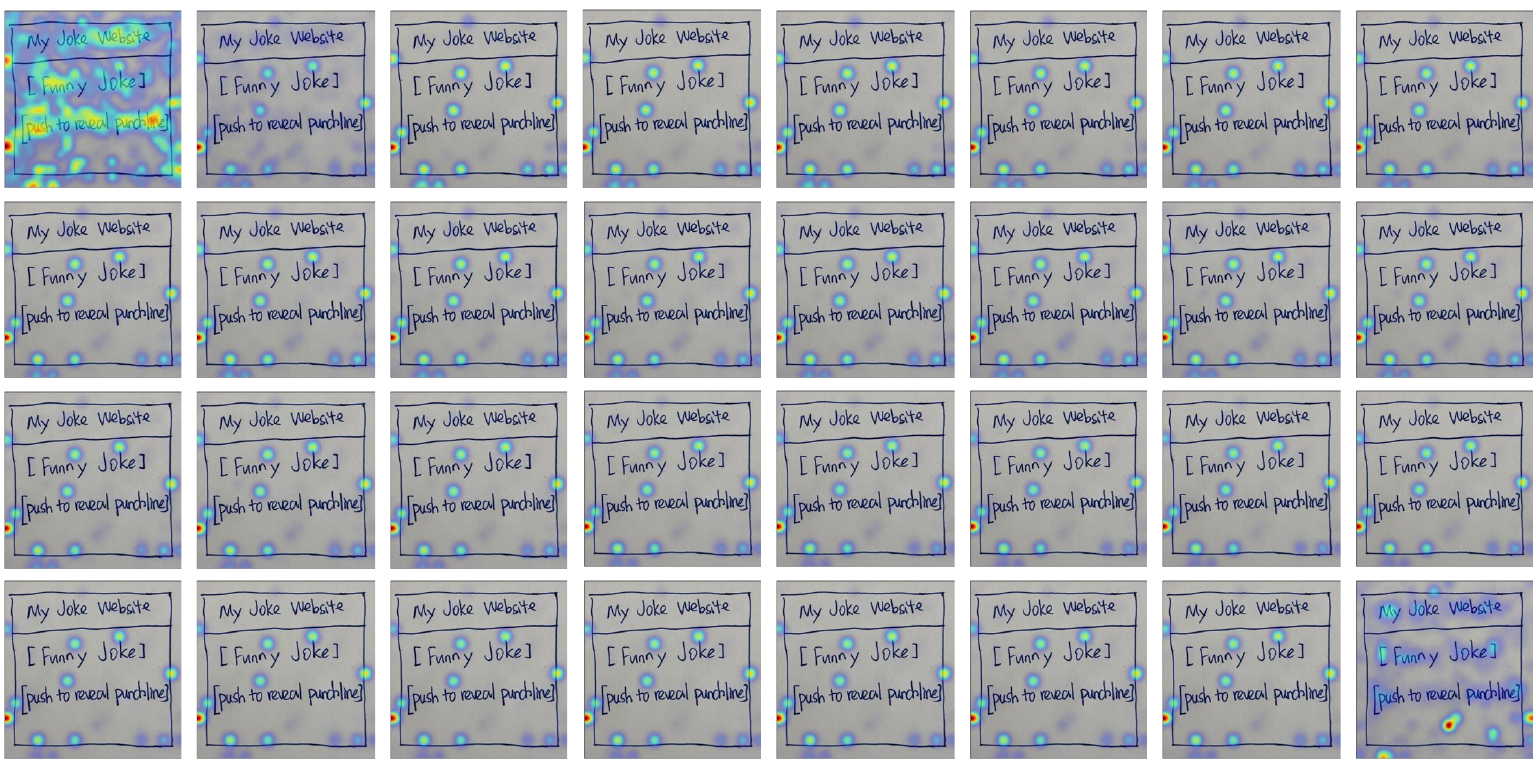}
            \subcaption{All-One}
        \end{minipage}

        \begin{minipage}[b]{.65\textwidth}
        \centering
            \includegraphics[width=\textwidth]{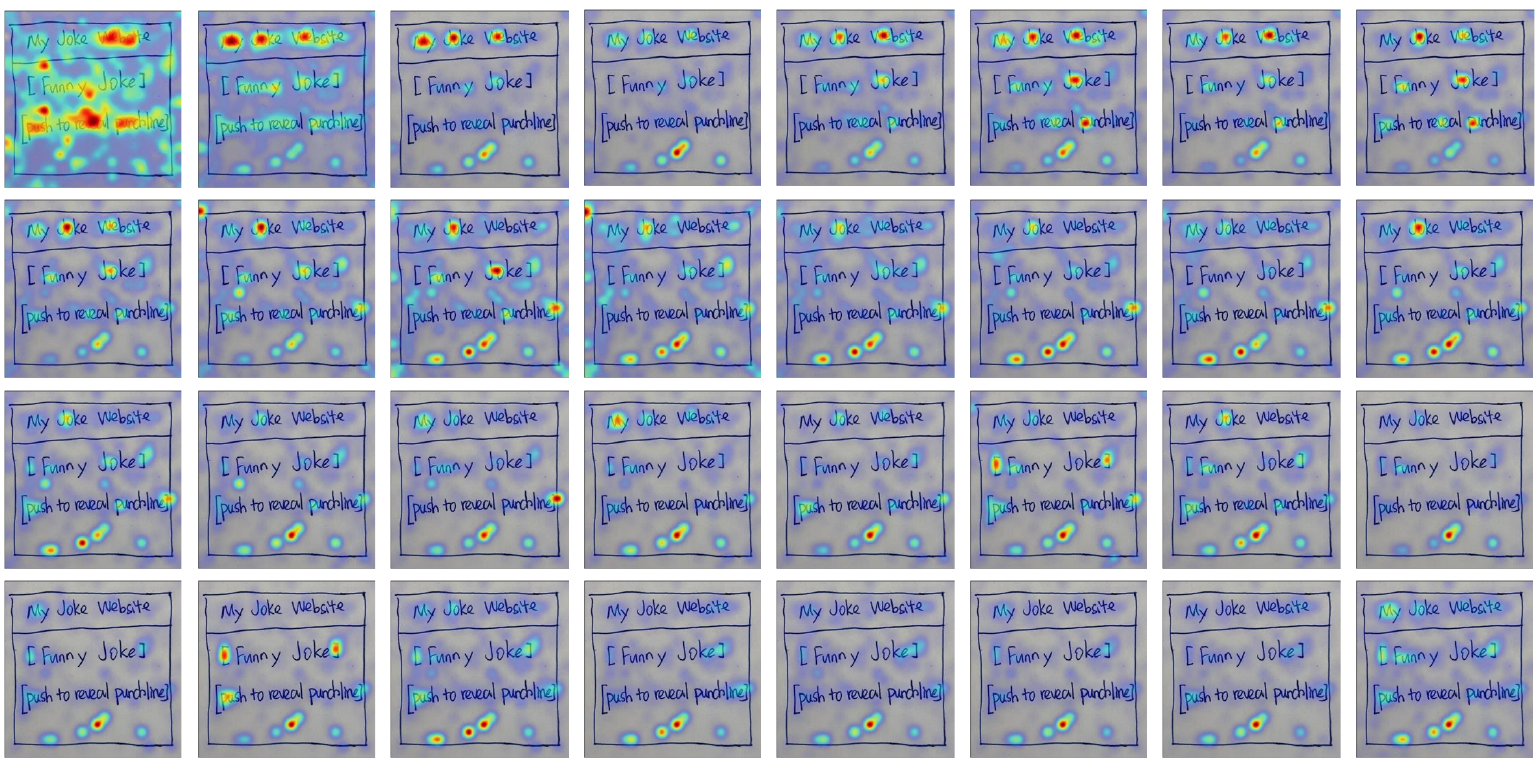}
            \subcaption{PyPE}
        \end{minipage}
    \caption{Layer-wise attention visualization (left to right, up to down) of the second example from Table~\ref{tab:more_cases}.}
    \label{fig:more_vis_2}
\end{figure*}

We provide more examples of visual description in Table~\ref{tab:more_cases}. As illustrated in the table, our proposed PyPE exhibits a reduced incidence of generating visual hallucinations or misunderstandings. More importantly, compared to other baseline methods, PyPE demonstrates a finer granularity in perceiving visual elements, thereby uncovering additional information, such as "blueberries" in the first example and "My joke website (funny joke push to reveal punchline)" in the second example. To further analyze the model's attention distribution across each decoder layer, we visualize the corresponding attention values for these examples. The results in Figure~\ref{fig:more_vis_0},~\ref{fig:more_vis_1}, and~\ref{fig:more_vis_2} indicate that while other baselines remain hindered by anchor tokens, PyPE consistently mitigates this issue, facilitating a more rational allocation of attention.

\begin{table*}[t!]
\centering
\setlength{\tabcolsep}{9pt} 
\resizebox{\linewidth}{!}{
\begin{tabular}{p{2cm} p{17.2cm} }
\toprule
 \multicolumn{2}{l}{\bf Visual input example, Visual Description 1}  \\
\midrule
&  \includegraphics[height=3cm]{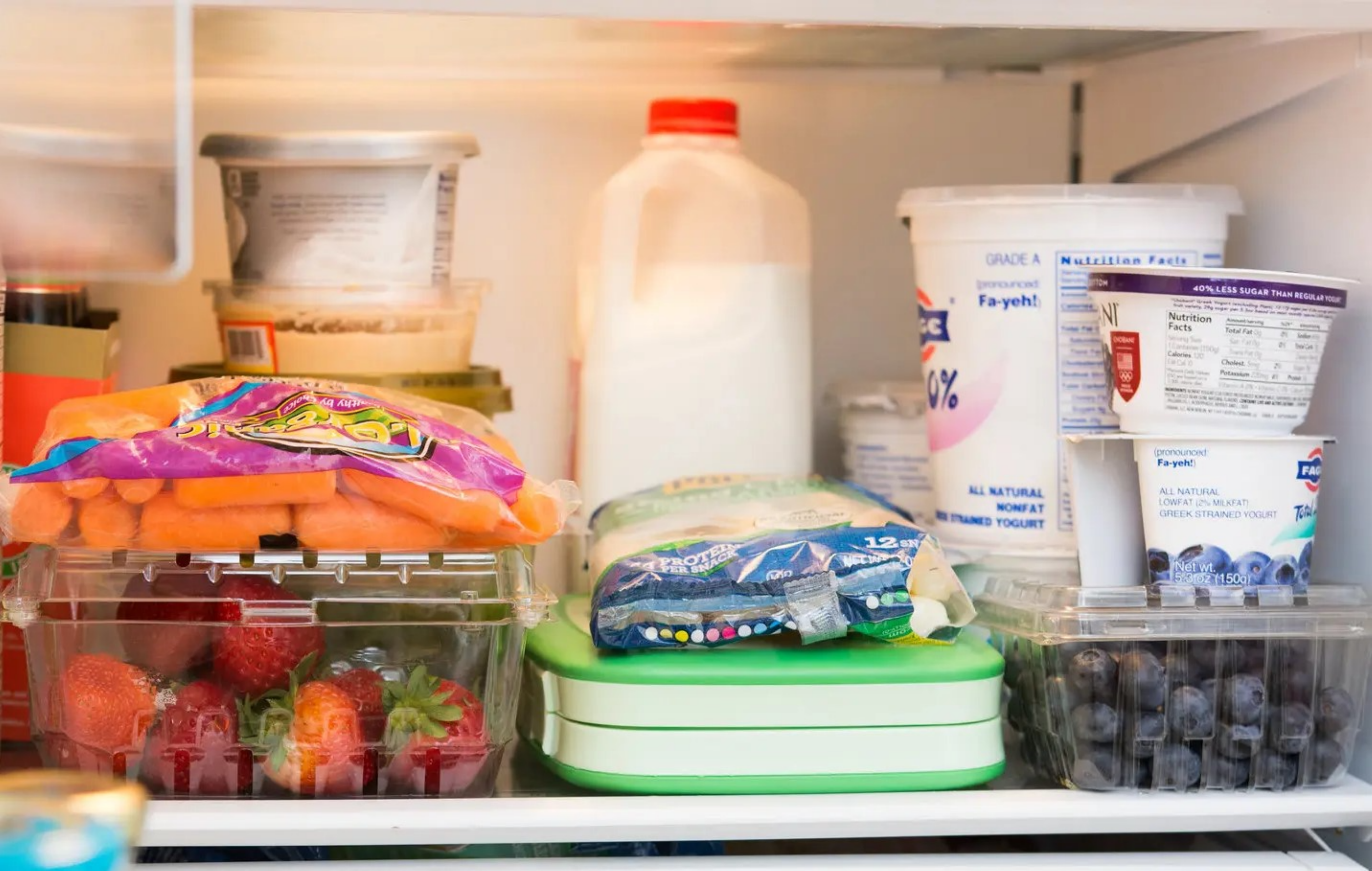} \\

User & Describe this photo in detail. \\
\midrule
Raster-scan & The image displays a refrigerator filled with various food items, including milk, juice, and fruits. There are \textbf{\textcolor{BrickRed}{multiple bottles of milk}} placed throughout the refrigerator, with some located \textbf{\textcolor{BrickRed}{on the top shelves and others on the bottom shelves}}. In addition to the milk, there are \textbf{\textcolor{BrickRed}{several bottles of juice, some of which are placed on the top shelves and others on the bottom shelves}}. A few carrots can be seen in the refrigerator, with some placed on the top shelves and others on the bottom shelves. The refrigerator is well-stocked with a variety of food items, making it a convenient place to store and access groceries. \\
\midrule
All-One & The image displays a refrigerator filled with various food items. Inside the refrigerator, there are several bottles, including \textbf{\textcolor{BrickRed}{one on the left side, two in the middle, and one on the right side}}. There are also multiple carrots, with some placed in the middle and others on the left side of the refrigerator. In addition to the carrots, there are several strawberries, with some located on the left side and others \textbf{\textcolor{BrickRed}{on the right side of the refrigerator.}} A bowl can be seen on the left side of the refrigerator, possibly containing more food items. The refrigerator is well-stocked with a variety of fruits and vegetables, making it a healthy and diverse selection. \\
\midrule
Concentric & The image displays a refrigerator filled with various food items, including a variety of fruits and vegetables. There are several carrots placed throughout the refrigerator, with some in the top left, middle, and bottom sections. A few strawberries can be seen in the bottom left corner of the refrigerator. In addition to the fresh produce, there are \textbf{\textcolor{BrickRed}{multiple bottles of milk and juice}} stored in the refrigerator. Some of these bottles are located in the \textbf{\textcolor{BrickRed}{top left, middle, and bottom}} sections of the refrigerator. The refrigerator is well-stocked with a diverse selection of food items, making it a convenient place to store and access groceries. \\
\midrule
\textbf{PyPE (ours)} &  The image displays a refrigerator filled with various food items, including several containers of berries, milk, and other foods. There are \textbf{\textcolor{BrickRed}{multiple bottles of milk}} placed throughout the refrigerator, with some located on the top shelves and others on the bottom shelves. In addition to the milk, there are several containers of berries, such as strawberries and \textbf{\textcolor{Green}{blueberries}}, spread across the refrigerator. The refrigerator is well-stocked with a diverse range of food items, making it a convenient place to store and access groceries. \\
\midrule
 \multicolumn{2}{l}{\bf Visual Description 2}  \\
\midrule
&  \includegraphics[height=3cm]{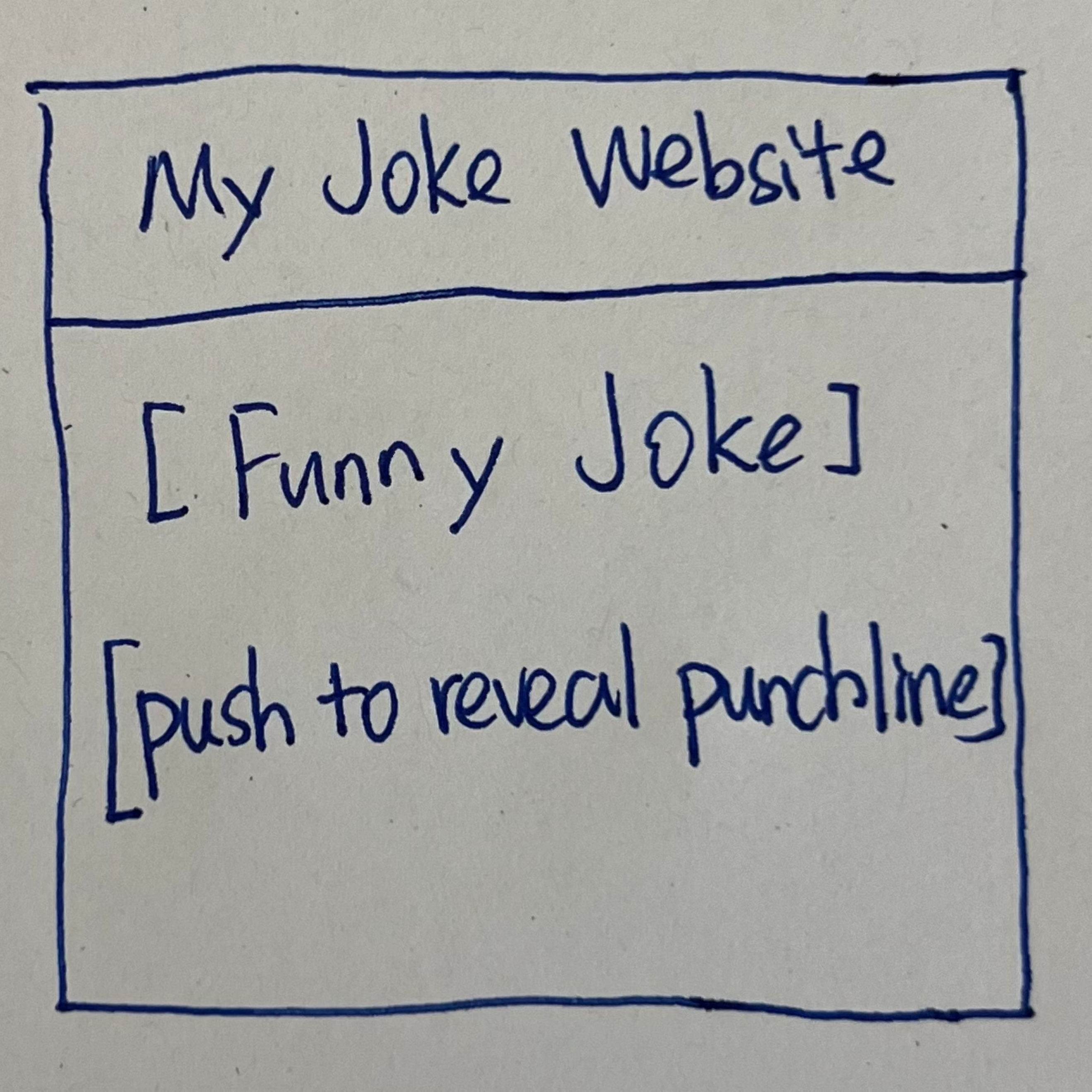} \\
User & Describe this photo in detail. \\
\midrule
Raster-scan & The image features a handwritten sign with a joke written on it. \textbf{\textcolor{BrickRed}{The sign is placed on a table}}, and it appears to be a website or a piece of paper with a humorous message. The joke is written in a \textbf{\textcolor{BrickRed}{foreign language}}, which adds a unique touch to the sign. The sign is placed in the center of the image, drawing attention to its content. \\
\midrule
All-One & The image features a handwritten sign with a joke website written on it. The sign is placed on a white surface, possibly a piece of paper or a wall. The joke website is written in blue ink, and \textbf{\textcolor{BrickRed}{the text is accompanied by a question mark, indicating that the joke is a riddle or a puzzle}}. The sign is placed in the center of the image, drawing attention to the joke website. \\
\midrule
Concentric & The image features a handwritten sign with a joke on it, likely meant to be displayed on a website. The joke is written in a \textbf{\textcolor{BrickRed}{foreign language, and the sign is accompanied by a question mark, indicating that it is meant to be a humorous or amusing statement}}. The sign is placed on a white background, which contrasts with the dark blue writing. \\
\midrule
\textbf{PyPE (ours)} &  The image features a handwritten sign with a joke on it, likely meant to be displayed on a website. The joke is written in blue ink, and it reads, \textbf{\textcolor{Green}{"My joke website (funny joke push to reveal punchline)."}} The sign is placed on a white background, which contrasts with the blue ink used for the joke. The handwriting style suggests a casual and humorous tone, likely meant to entertain or amuse the viewers. \\
\bottomrule
\end{tabular}
}
\captionof{table}{More examples from LLaVA-Bench. The misunderstandings and hallucinations of visual contents are highlighted in \textbf{\textcolor{BrickRed}{red}}. The descriptions that are not mentioned in baselines but are accurately represented by PyPE are highlighted in \textbf{\textcolor{Green}{green}}.}
\label{tab:more_cases}
\end{table*}

\end{document}